\documentclass{article}

\PassOptionsToPackage{numbers, compress, sort}{natbib}

\usepackage[preprint]{neurips_2020}

\usepackage[utf8]{inputenc} 
\usepackage[T1]{fontenc}    
\usepackage{url}            
\usepackage{booktabs}       
\usepackage{xcolor}
\usepackage{float}
\usepackage{graphicx}
\usepackage{amsmath}
\usepackage{bm}
\usepackage{amsfonts}       
\usepackage{amssymb}
\usepackage{nicefrac}       
\usepackage{microtype}      
\usepackage{enumitem}
\usepackage{multirow}
\usepackage{array}
\usepackage[ruled,vlined]{algorithm2e}
\usepackage{url}

\usepackage{breakurl}
\usepackage[breaklinks]{hyperref}
\hypersetup{colorlinks = true, citecolor = red, linkcolor = blue, urlcolor = black}       
\usepackage{bookmark}
\usepackage[font = small]{caption}

\newcolumntype{L}[1]{>{\raggedright\let\newline\\\arraybackslash\hspace{0pt}}m{#1}}
\newcolumntype{C}[1]{>{\centering\let\newline\\\arraybackslash\hspace{0pt}}m{#1}}
\newcolumntype{R}[1]{>{\raggedleft\let\newline\\\arraybackslash\hspace{0pt}}m{#1}}

\makeatletter
\newcommand{\settitle}{\@maketitle}
\makeatother

\newcommand\blfootnote[1]{%
	\begingroup
	\renewcommand\thefootnote{}\footnote{#1}%
	\addtocounter{footnote}{-1}%
	\endgroup
}

\newcommand{\QQ}{{\mathbb Q}}
\newcommand{\PP}{{\mathbb P}}
\newcommand{\Ex}{{\mathbb E}}

\newcommand{\BlackBox}{\rule{1.5ex}{1.5ex}}  

\newtheorem{theorem}{Theorem}
 
\newtheorem{proposition}{Proposition} 

\newtheorem{corollary}{Corollary}
\newtheorem{definition}{Definition}

\title{$f$-Divergence Variational Inference}

\author{
	Neng Wan$^{1}$\thanks{\ Authors contributed equally to this paper.}\\
	\texttt{nengwan2@illinois.edu} \\
	\And
	Dapeng Li$^{\hspace{1pt}2\hspace{1pt}*}$\\
	\texttt{dapeng.ustc@gmail.com} \\
	\And
	Naira Hovakimyan$^1$ \\
	\texttt{nhovakim@illinois.edu} \\
}

\begin{document}
	
	\maketitle
	\vspace{-30pt}
	\begin{center}
		$^{1}$ University of Illinois at Urbana-Champaign, Urbana, IL 61801 \\
		$^{2}$ Anker Innovations, Shenzhen, China\\
	\end{center}
	\vspace{30pt}
	
	\begin{abstract}
		This paper introduces the \textit{$f$-divergence variational inference} ($f$-VI) that generalizes variational inference to all $f$-divergences. Initiated from minimizing a crafty surrogate $f$-divergence that shares the statistical consistency with the $f$-divergence, the $f$-VI framework not only unifies a number of existing VI methods, \textit{e.g.} Kullback–Leibler VI~\cite{Jordan_ML_1999}, R\'{e}nyi's $\alpha$-VI~\cite{Li_NIPS_2016}, and $\chi$-VI~\cite{Dieng_NIPS_2017}, but offers a standardized toolkit for VI subject to arbitrary divergences from $f$-divergence family. A general $f$-variational bound is derived and provides a sandwich estimate of marginal likelihood (or evidence). The development of the $f$-VI unfolds with a stochastic optimization scheme that utilizes the reparameterization trick, importance weighting and Monte Carlo approximation; a mean-field approximation scheme that generalizes the well-known coordinate ascent variational inference (CAVI) is also proposed for $f$-VI. Empirical examples, including variational autoencoders and Bayesian neural networks, are provided to demonstrate the effectiveness and the wide applicability of $f$-VI. 
	\end{abstract}

	\section{Introduction}
	Variational inference (VI) is a machine learning method that makes Bayesian inference computationally efficient and scalable to large datasets. For decades, the dominant paradigm for approximate Bayesian inference $p(z|x) = p(z, x) / p(x)$ has been Markov-Chain Monte-Carlo (MCMC) algorithms, which estimate the evidence $p(x) = \int p(z, x) dz$ via sampling. However, since sampling tends to be a slow and computationally intensive process, these sampling-based approximate inference methods fade when dealing with the modern probabilistic machine learning problems that usually involve very complex models, high-dimensional feature spaces and large datasets. In these instances, VI becomes a good alternative to perform Bayesian inference. The foundation of VI is primarily optimization rather than sampling. To perform VI, we posit as a family of approximate (or recognition) densities $\mathcal{Q}$ and find the member $q^*(z) \in \mathcal{Q}$ that minimizes the statistical divergence to the true posterior $D(q(z)\|p(z|x))$. Meanwhile, since VI also has many elegant and favorable theoretical properties, \textit{e.g.} variational bounds of the true evidence, it has become the foundation of many popular generative and machine learning models.  

	Recent advances in VI can be roughly categorized into three groups, improvements over traditional VI algorithms~\cite{Knowles_NIPS_2011, Wang_JMLR_2013}, developments of scalable VI methods~\cite{Hoffman_JMLR_2013, Kingma_ICLR_2014, Li_NIPS_2015}, and explorations for tighter variational bounds~\cite{Burda_ICLR_2016, Tao_ICML_2018}. Comprehensive reviews on VI's background and progression can be found in~\cite{Blei_JASA_2017, Zhang_TPAMI_2019}. While most of these advancements were built on the classical VI associated with the Kullback–Leibler (KL) divergence, some recent efforts tried to extend the VI framework to other statistical divergences and showed promising results. Among these efforts, R\'enyi's $\alpha$-divergence and $\chi$-divergence as the \textit{root} divergences (or generators) of the KL divergence were employed for VI in~\cite{Li_NIPS_2016, Dieng_NIPS_2017, Regli_arXiv_2018}, which not only broadens the variety of statistical divergences for VI, but makes KL-VI a special case of their methods. Stochastic optimization methods from KL-VI, such as stochastic VI~\cite{Hoffman_JMLR_2013} and black-box VI~\cite{Ranganath_ICAIS_2014}, were generalized to R\'enyi's $\alpha$-VI and $\chi$-VI in~\cite{Li_NIPS_2016, Dieng_NIPS_2017}, and the modified algorithms with new divergences outperformed the classical KL-VI in some benchmarks of Bayesian regressions and image reconstruction. Nevertheless, mean-field approximation, an important type of KL-VI algorithms including the coordinate ascent variational inference (CAVI) and expectation propagation (EP) algorithms~\cite{Bishop_2006, Minka_UAI_2001, Blei_JASA_2017}, were regretfully not revisited or extended for these new divergences.

	As the root divergence of the R\'enyi's $\alpha$-divergence, $\chi$-divergence and many other useful divergences~\cite{Sason_TIT_2016, Sason_Entropy_2018}, $f$-divergence is a more inclusive statistical divergence (family) and was utilized to improve the statistical properties~\cite{Bamler_NIPS_2017, Wang_NIPS_2018}, sharpness~\cite{Tao_ICML_2018, Zhang_arxiv_2019}, and surely the generality of variational bounds~\cite{Tao_ICML_2018, Zhang_arxiv_2019, Knoblauch_arXiv_2019}. However, most of these works only dealt with some portions of $f$-divergences for their favorable statistical properties, \textit{e.g.} mass-covering~\cite{Bamler_NIPS_2017} and tail-adaptive~\cite{Wang_NIPS_2018}, and did not develop a systematic VI framework that harbors all $f$-divergences. Meanwhile, since i) the regular $f$-divergence does not explicitly induce an $f$-variational bound as elegant as the ELBO~\cite{Blei_JASA_2017}, $\chi$ upper bound (CUBO)~\cite{Dieng_NIPS_2017}, or R\'enyi variational bound (RVB)~\cite{Li_NIPS_2016}, and ii) only specific choices of $f$-divergence result in an $f$-variational bound that trivially depends on the evidence~\cite{Zhang_TPAMI_2019}, a thorough and comprehensive analysis on the $f$-divergence VI has been due for a long time.

	In this paper, we extend the traditional VI to $f$-divergence, a rich family that comprises many well-known divergences as special cases~\cite{Sason_TIT_2016}, by offering some new insights into the $f$-divergence VI and a unified $f$-VI framework that encompasses a number of recent developments in VI methods. An explicit benefit of $f$-VI is that it allows to perform VI or Bayesian approximation with even more variety of divergences, which can potentially bring us sharper variational bounds, more accurate estimate of true evidence, faster convergence rates, more criteria for selecting approximate model $q(z)$, \textit{etc}. We hope this effort can be the last brick to complete the building of $f$-divergence VI and motivate more useful and efficient VI algorithms in the future. After reviewing the $f$-divergence and introducing a crafty surrogate $f$-divergence that is interchangeable with the regular $f$-divergence, we make the following contributions:
	\begin{enumerate}[label=$c$\arabic*), leftmargin=2em, topsep=-3pt, itemsep = 3pt, partopsep = 0pt, parsep = 0pt]
		\item We enrich the $f$-divergence VI theory by introducing an $f$-VI scheme via minimizing a surrogate $f$-divergence, which makes our $f$-VI framework compatible with the traditional VI approaches and naturally unifies an amount of existing VI methods, such as KL-VI~\cite{Jordan_ML_1999}, $\alpha$-VI~\cite{Li_NIPS_2016}, $\chi$-VI~\cite{Dieng_NIPS_2017}, and their related developments~\cite{Kingma_ICLR_2014, Burda_ICLR_2016, Wang_NIPS_2018, Tao_ICML_2018, Li_NIPS_2015}. 
		\item  We derive an $f$-variational bound for the evidence and equip it with the upper/lower bound criteria and an importance-weighted (IW-)bound. The $f$-variational bound is realized with an arbitrary $f$-divergence and unifies many existing bounds, such as ELBO, CUBO, RVB, and a number of generalized evidence bounds (GLBO)~\cite{Tao_ICML_2018}. 	  
		\item  We propose a universal optimization solution that comprises a stochastic optimization algorithm and a mean-field approximation algorithm for $f$-VI subject to all $f$-divergences, whether or not the $f$-variational bounds trivially depend on the evidence. Experiments on Bayesian neural networks and variational autoencoders (VAEs) show that $f$-VI can be comparable to, or even better than, a number of the state-of-the-art variational methods.
	\end{enumerate}

	\vspace{-0.2em}
	\section{Preliminary of $f$-divergence}
	\vspace{-0.2em}
	We first introduce some definitions and properties related to $f$-divergence, which are to be adopted in our subsequent exposition.
	
	\vspace{-0.2em}
	\subsection{$f$-divergence}\vspace{-0.2em}
	An $f$-divergence that measures the difference between two continuous probability distributions $q$ and $p$ can be defined as follows~\cite{Sason_TIT_2016}.
	\begin{definition}\label{def1}
		The $f$-divergence from  probability density functions $q(z)$ to $p(z)$ is defined as 
		\begin{equation}\label{F_Div}
			D_f(q(z) \| p(z) ) =:  \int f \left (\frac{q(z) }{p(z) }  \right )  p(z) \ dz = \Ex_{p} \left[f \left(  \frac{q(z)}{p(z)}\right) \right],
		\end{equation}
		where $f(\cdot)$ is a convex function with $f(1)=0$.
	\end{definition}
	\hyperref[def1]{Definition~1} assumes that $q(z)$ is absolutely continuous w.r.t. $p(z)$, which might not be exhaustive, but avoids the unnecessary entanglements with measure theory details. One can however refer to~\cite{Sason_TIT_2016, Sason_Entropy_2018} for a more rigorous treatment. Most prevailing divergences adopted in VI can be regarded as the special cases of $f$-divergence and hence be restored by choosing a proper $f$-function $f(\cdot)$. \hyperref[tab1]{Table~1} and~\cite{Sason_TIT_2016, Sason_Entropy_2018, Zhang_arxiv_2019} present the relationship between some well-known statistical divergences adopted in VI and their $f$-functions. Intuitively, one can perform $f$-VI by minimizing either the \textit{forward} $f$-divergence $D_f(p\|q)$ or the \textit{reverse} $f$-divergence $D_f(q\|p)$, and~\cite{Murphy_2012, Zhang_arxiv_2019} provide some heuristic discussions on their statistical differences. Since VI based on the reverse KL divergence is more tractable to compute and more statistically sensible, we will develop our $f$-VI framework primarily based on the reverse $f$-divergence, while one can still unify or commute between the forward and reverse $f$-divergences via the \textit{dual function} $f^*$, which is also referred to as the perspective function or the conjugate symmetry of $f$ in~\cite{Boyd_2004, Sason_TIT_2016, Dieng_NIPS_2017}.
	\begin{definition}
		Given a function $f: (0, \infty) \rightarrow \mathbb{R}$, the dual function $f^*: (0, \infty) \rightarrow \mathbb{R}$ is defined as 
		\begin{equation*}
			f^*(t)= t \cdot f(1/t).
		\end{equation*}
	\end{definition}
	One can verify that the dual function $f^*$ has the following properties: i) $(f^*)^* = f$; ii) if $f$ is convex, $f^*$ is also convex, and iii) if $f(1) = 0$, then $f^*(1) = 0$. With dual function $f^*$, an identity between the forward and reverse $f$-divergences can be established~\cite{Dieng_NIPS_2017}:
	\begin{equation*}
		D_{f^*}(p \| q) = \int \frac{p(z)}{q(z)} \cdot f\left(\frac{q(z)}{p(z)}\right) \cdot q(z) \ dz  = D_f(q \| p).
	\end{equation*}

	In order to facilitate the derivation of $f$-variational bound, especially when the latent variable model is involved~\cite{Nowozin_NIPS_2016, Zhang_arxiv_2019}, we introduce a \textit{surrogate $f$-divergence} $D_{f^{}_\lambda}$ defined by the \textit{generator function} 
	\begin{equation}\label{Generator_Fun}
		f_{\lambda}(\cdot) = f( \lambda \cdot)- f(\lambda),
	\end{equation}
	where  $\lambda \geq 0$ is constant. It is straightforward to verify that $f$ and $f_\lambda$ have the same convexity, and $f(1) = 0$ implies $f_\lambda(1) = 0$, which induces a valid (surrogate) $f$-divergence, denoted as $D_{f_\lambda}$, that can virtually replace $D_f$ when needed\footnote{Essentially, $D_{f^{}_{\lambda}} $ is an $f$-divergence between a positive measure $\PP(\cdot, \lambda) $ and a probability measure $\QQ(\cdot )$.}. To justify the closeness between divergences $D_f$ and $D_{f_\lambda}$, we first note that $D_f$ and $D_{f_\lambda}$ share the same minimum point at $p = q$, then  we have the following statement.
	\begin{proposition}\label{prop1}
		Given two probability distributions $q$ and $p$, a convergent sequence $\lim_{n \rightarrow \infty} \lambda_n = 1, \lambda_n\geq0$, and a convex function $f: (0, +\infty) \rightarrow \mathbb{R}$ such that $f(1) = 0$ and $f(\cdot)$ is uniformly continuous, the $f$-divergences between $q$ and $p$ satisfy
		\begin{equation}\label{prop1_eq1}
			D_{f^{}_{\lambda_n}}(q \| p)  \rightarrow D_{f}(q \| p)
		\end{equation}
		almost everywhere as $n\rightarrow \infty$.
	\end{proposition}

	\vspace{-0.2em}
	\subsection{Shifted homogeneity}\vspace{-0.2em}
	We then introduce a class of $f$-functions equipped with a structural advantage in decomposition, which will be invoked later to derive the coordinate-wise VI algorithm under mean-field assumption.
	\begin{definition}
		A convex function $f$ belongs to $\mathcal{F}_{\{0, 1\}}$, if $f(1) = 0$, and for any $t,\tilde t \in \mathbb R$, we have
		\begin{equation}\label{ShiftHomo}
			f( t \tilde{t} )=  t^\gamma f(\tilde t)+f( t ) {\tilde t}^\eta \,,
		\end{equation}
		where $\gamma \in \mathbb{R}$, and $\eta \in \{0,1\}$. Function $f$ is type $0$ shifted homogeneous or $f\in\mathcal{F}_0$ if $\eta = 0$, and type $1$ shifted homogeneous or $f \in \mathcal{F}_1$ if $\eta = 1$.
	\end{definition}
	This special class of functions allows to decompose an $f$-function into two or more (by iterations) terms, each of which is a product of an $f$-function and an exponent.  In \hyperref[tab1]{Table~1}, we show that the $f$-functions of many well-known divergences can be classified as $\mathcal{F}_{\{0, 1\}}$ functions.\vspace{-10pt}
	\begin{table}[H]
		\caption{Divergences $D_f^{}(q\|p)$ and homogeneity decomposition.}\vspace{0.2em}
		\label{tab1}
		\centering\small
		\begin{tabular}{lcc}
			\toprule[1pt]
			Divergences     & $f(t)$     &  $f(t \tilde{t})$ \\
			\midrule[0.3pt]
			KL divergence~\cite{Jordan_ML_1999} & $t \log t$ & $ t f(\tilde t) + f(t) \tilde{t}$\\
			General $\chi^n$-divergence~\cite{Dieng_NIPS_2017} & $t^n -1, n \in \mathbb{R} \backslash (0, 1)$ & $t^n f(\tilde{t}) + f(t)$\\
			Hellinger $\alpha$-divergence $\mathcal{H}_\alpha^{}$~\cite{Sason_Entropy_2018}  & $(t^{\alpha} - 1) / (\alpha - 1), \alpha \in \mathbb{R}^{+} \backslash \{1\}$  & $t^\alpha f(\tilde{t}) + f(t)$\\
			R\'{e}nyi's $\alpha$-divergence\protect\footnotemark~\cite{Li_NIPS_2016} & \multicolumn{2}{c}{$D_\alpha(q\|p) = (\alpha - 1)^{-1}  \log[1+(\alpha - 1)\mathcal{H}_\alpha(q\|p)]$}\\
			\bottomrule[1pt]
		\end{tabular}
	\end{table}\footnotetext{Renyi's $\alpha$-divergence cannot be directly restored from $f$-divergence~\eqref{F_Div}, while it is a one-to-one transformation of $\mathcal{H}_\alpha$ of the same order $\alpha \in \mathbb{R}^{+} \backslash \{1\}$.}\vspace{-5pt}
	The duality property between $\mathcal F_0$ and $\mathcal F_1$ is stated in~\hyperref[prop2]{Proposition~2}. 
	\begin{proposition}\label{prop2}
		Given $f_0 \in \mathcal F_0$ and  $f_1 \in \mathcal F_1$, the dual functions $f_0^*\in \mathcal F_1$ and $f_1^* \in \mathcal F_0$.
	\end{proposition}
	When $f\in\mathcal{F}_{\{0, 1\}}$, we can establish a more profound relationship, in contrast with \hyperref[prop1]{Proposition~1}, between $f$-divergence $D_f$ and surrogate divergence $D_{f_\lambda}$.
	\begin{proposition}\label{prop3}
		When $f\in \mathcal F_{\{0, 1\}}$ and $\lambda > 0,$ an $f$-divergence $D_f$ and its surrogate divergence $D_{f_\lambda}$ satisfy 
		\begin{equation}\label{thm1eq}
			D_{f_\lambda}(q \| p) = \lambda^\gamma D_f(q \| p).
		\end{equation}
	\end{proposition}
	By virtue of the equivalence relationship revealed in~\hyperref[prop1]{Proposition~1} and~\hyperref[prop3]{3}, we can interchangeably use $f$-divergence $D_f$ and surrogate divergence $D_{f_\lambda}$, and the parameter $\lambda$ of surrogate divergence provides an additional degree of freedom when deriving the variational bounds and VI algorithms.

	\section{Variational bounds and optimization}
	While it was difficult to retrieve an $f$-variational bound~\cite{Tao_ICML_2018, Wang_NIPS_2018, Zhang_arxiv_2019}, which is an expectation over $q$ and unifies the existing variational bounds~\cite{Blei_JASA_2017, Li_NIPS_2016, Dieng_NIPS_2017}, by directly manipulating the original $f$-divergence in~\eqref{F_Div}, we will show that such a general variational bound can be found when minimizing a crafty surrogate $f$-divergence.

	\subsection{$f$-variational bounds}\label{sec3_1}

	Given a convex function $f$ such that $f(1) = 0$ and a set of i.i.d. samples $\mathcal{D} = \{{x}^{(n)}\}_{n=1}^N$, the generator function $f_{p(\mathcal{D})^{-1}}$ with $p(\mathcal{D}) > 0$ can induce a surrogate $f$-divergence. Our $f$-VI is then initiated from minimizing the following reverse (surrogate) $f$-divergence
	\begin{equation}\label{VB1}
		D_{f_{p(\mathcal{D})^{-1}}}\left (q(z) \| p(z|\mathcal{D})\right ) = \frac{1}{p(\mathcal{D})} \cdot \Ex_{q(z)} \left[ f^*\left ( \frac{p(z, \mathcal{D})}{q(z)}\right ) \right] - f\left ( \frac{1}{p(\mathcal{D})} \right ).
	\end{equation} 
	Multiplying both sides of~\eqref{VB1} by $p(\mathcal{D})$ and with rearrangements, we have
	\begin{equation}\label{VB2}
		\mathcal{L}_f(q, \mathcal{D}) = \Ex_{q(z)} \left[ f^*\left ( \frac{p(z, \mathcal{D})}{q(z)}\right )  \right]  =  f^*(p(\mathcal{D})) + p(\mathcal{D}) \cdot D_{p(\mathcal{D})^{-1}}\left (q(z) \| p(z|\mathcal{D})\right ). 
	\end{equation}
	For a given evidence $p(\mathcal{D})$, we can minimize the $f$-divergence $D_{f_{p(\mathcal{D})^{-1}}}\left (q(z) \| p(z|\mathcal{D})\right )$ by minimizing the expectation in~\eqref{VB2}, which is defined as the $f$-variational bound $\mathcal{L}_f(q, \mathcal{D})$. Consequently, by the non-negativity of $f$-divergence~\cite{Sason_TIT_2016, Sason_Entropy_2018}, we can establish the following inequality.
	\begin{theorem}\label{thm1}
		Dual function of evidence $f^*(p(\mathcal{D}))$ is bounded above by $f$-variational bound $\mathcal{L}_f(q, \mathcal{D})$
		\begin{equation}\label{f_bound}
			\mathcal{L}_{f}(q, \mathcal{D}) = \Ex_{q(z)} \left[ f^*\left( \frac{p(z, \mathcal{D})}{q(z)} \right) \right] \geq f^*(p(\mathcal{D})),
		\end{equation}
		and equality is attained when $q(z)=p(z|\mathcal{D})$, \textit{i.e.} $D_{p(\mathcal{D})^{-1}}\left(q(z) \| p(z|\mathcal{D})\right) = 0$.\footnote{Inequality~\eqref{f_bound} can also be derived by resorting to Jensen's inequality. Since $f^*$ is convex, we have 
			\begin{equation*}
				\Ex_{q(z)} \left[ f^*\left( \frac{p(z,\mathcal{D})}{q(z)} \right) \right]     \geq f^*\left( \Ex_{q(z)} \left[  \frac{p(z, \mathcal{D})}{q(z)}  \right] \right)= f^*(p(\mathcal{D})).
		\end{equation*}}
	\end{theorem}
	By properly choosing $f$-function, $f$-variational bound $\mathcal{L}_f(q, \mathcal{D})$ and~\eqref{f_bound} can restore the most existing variational bounds and the corresponding inequalities, \textit{e.g.} $f(t) = t\log(t)$ for ELBO in~\cite{Blei_JASA_2017} and $f(t) = t^{1-n}-t$ for CUBO in~\cite{Dieng_NIPS_2017}. See Supplementary Material (SM) for more restoration examples and some new variational bounds, \textit{e.g.} an evidence upper bound (EUBO) under KL divergence. While the assumption of $p(\mathcal{D}) > 0$ or the existence of $p(\mathcal{D})^{-1}$ in~\eqref{VB1} might lay additional restrictions in some situations, we can circumvent them by resorting to the $f$-VI minimizing the forward surrogate $f$-divergence $D_{f_{p(\mathcal{D})}}(p(z|\mathcal{D})\|q(z))$. SM provides more details for this alternative. Additionally, $\mathcal{L}_{f}(q, \mathcal{D})$ in~\eqref{f_bound} can be further sharpened by leveraging multiply-weighted posterior samples~\cite{Burda_ICLR_2016}, \textit{i.e.}, importance-weighted VI.
	\begin{corollary}\label{cor0}
		When $1 \leq L_1 \leq L_2$, the importance-weighted $f$-variational bound $\mathcal{L}^{\text{\rm IW}}_{f}(q, \mathcal{D}, L)$ and the $f$-variational bound $\mathcal{L}_{f}(q, \mathcal{D})$ satisfy  
		\begin{equation*}
			\mathcal{L}_{f}(q, \mathcal{D}) \geq \mathcal{L}^{\text{\rm IW}}_{f}(q, \mathcal{D}, L_1) \geq \mathcal{L}^{\text{\rm IW}}_{f}(q, \mathcal{D}, L_2)  \xrightarrow{L\rightarrow \infty} f^*(p(\mathcal{D})),
		\end{equation*}
		where $\mathcal{L}^{\text{\rm IW}}_{f}(q, \mathcal{D}, L)$ is defined as \vspace{-3pt}
		\begin{equation*}
			\mathcal{L}^{\text{\rm IW}}_{f}(q, \mathcal{D}, L) = \Ex_{z_{1:L} \sim q(z)} \left[   f^*\left(  \frac{1}{L} \sum_{l=1}^{L} \frac{p(z_l, \mathcal{D})}{q(z_l)} \right) \right],
		\end{equation*}
		and $z_{1:L} = \{z_l\}_{l=1}^L$ are $L \in \mathbb{N}^{*}$ i.i.d. samples from $q(z)$.
	\end{corollary}
	For clarity and conciseness, we will develop the subsequent results primarily based on $\mathcal{L}_{f}(q, \mathcal{D})$. Nevertheless, our readers should feel safe to replace $\mathcal{L}_{f}(q, \mathcal{D})$ with $\mathcal{L}^{\textrm{IW}}_{f}(q, \mathcal{D}, L)$ in the following context and obtain improved outcomes. More interesting results can be observed from~\eqref{f_bound}. After composing both sides of~\eqref{f_bound} with the inverse dual function $(f^*)^{-1}$, we have the following observations:
	\begin{enumerate}[label=$o$\arabic*), leftmargin=2em, topsep=0pt, itemsep = 0pt, partopsep = 0pt, parsep = 0pt]
		\item When the dual function $f^*$ is increasing (or non-decreasing) on $\mathbb R^+$, the composition gives an evidence upper bound:
		\begin{equation*}
			(f^*)^{-1}\circ \mathcal{L}_{f}(q, \mathcal{D})  \geq p(\mathcal{D}).
		\end{equation*}
		\item  When the dual function $f^*$ is decreasing (or non-increasing) on $\mathbb{R}^{+}$, the composition gives an evidence lower bound:
		\begin{equation*}
			(f^*)^{-1}\circ \mathcal{L}_{f}(q, \mathcal{D})   \leq p(\mathcal{D}).
		\end{equation*}
		\item  When the dual function $f^*$ is non-monotonic on $\mathbb{R}^{+}$, the composition gives a local evidence bound by applying $o1)$ or $o2)$ on a monotonic interval of $f^*$. 
	\end{enumerate}
	Based on these observations, we can readily imply a sandwich formula for evidence $p(\mathcal{D})$, which is essential for accurate VI~\cite{Zhang_TPAMI_2019}.
	\begin{corollary}\label{cor1}
		Given convex functions $f$ and $g$ such that $f(1)=g(1)=0$,  on an interval where $f^*$ is increasing and $g^*$ is decreasing, the evidence $p(\mathcal{D})$ satisfies
		\begin{equation}\label{Sandwich_Bound}
			(g^*)^{-1}\circ \Ex_{q(z)} \left[ g^* \left ( \frac{p(z,\mathcal{D})}{q(z)}\right ) \right]  \leq p(\mathcal{D}) \leq (f^*)^{-1}\circ \Ex_{q(z)} \left[ f^* \left( \frac{p(z, \mathcal{D})}{q(z)} \right) \right].
		\end{equation}	
	\end{corollary}
	The evidence bounds in~\eqref{Sandwich_Bound} are akin to the GLBO, which was proposed on the basis of a few assumptions and intuitions in~\cite{Tao_ICML_2018}. \hyperref[cor0]{Corollary~1} and \hyperref[cor1]{Corollary~2} interprets and supplements GLBO with rigorous $f$-VI analysis and explicit instructions on choosing $f$-function. Compared with the unilateral variational bounds, the bilateral bounds in~\eqref{Sandwich_Bound} reveal more information and allow to estimate $p(\mathcal{D})$ with more accuracy. To sharpen these bilateral bounds, we need to properly choose the functions $f$ and $g$ and the recognition model $q(z)$ such that $\sup_{g, q}{g}^{-1}\circ \mathcal{L}_{g}(q, \mathcal{D})$ and $\inf_{f, q}{f}^{-1}\circ \mathcal{L}_{f}(q, \mathcal{D})$ can be attained. For a selected family of $q(z)$, various choices of $f$ and $g$ will lead to evidence bounds of different sharpness and optimization efficiency. The model selection of approximate distribution $q(z)$ is a fundamental problem inherited by all VI algorithms, and a feasible solution is to compare the performance of candidate models while fixing an $f$- or $g$-function~\cite{Tao_ICML_2018} or alternating among some common divergences. Once the functions $f$ and $g$ and the recognition model $q(z)$ are determined, we can approximate the optimal distribution $q^*(z)$ in $q(z)$ or minimize $\mathcal{L}_{f}(q, \mathcal{D})$ by adjusting the parameters in $q(z)$, which does not require the dual function $f^*$ or $g^*$ be invertible as in~\eqref{Sandwich_Bound} and will be discussed in the succeeding subsections.

	\subsection{Stochastic optimization}\label{sec3_2}
	While classical VI is limited to conditionally conjugate exponential family models~\cite{Murphy_2012,  Blei_JASA_2017, Zhang_TPAMI_2019}, the stochastic optimization makes VI applicable for more modern and complicated problems~\cite{Hoffman_JMLR_2013, Ranganath_ICAIS_2014}. To minimize $\mathcal{L}_f(q, \mathcal{D})$ with stochastic optimization, we supplement the preceding VI formulation with more details. The approximate model is formulated as $q^{}_\theta(z)$, where $\theta \in \mathbb{R}^M$ are the parameters to be optimized. While some papers~\cite{Mnih_ICML_2014, Kingma_ICLR_2014, Tao_ICML_2018} also consider and optimize the parameters $\phi$ in the generative model $p_\phi$, we prefer to treat the parameters $\phi$ as latent variables $z$ for conciseness. An intuitive approach to apply stochastic optimization is to compute the standard gradient of $\mathcal{L}_f(q, \mathcal{D})$ or $\mathcal{L}^{\textrm{IW}}_f(q, \mathcal{D})$ w.r.t. $\theta$
	\begin{equation}\label{score_fun_gradient}
		\nabla^{}_{\theta} \mathcal{L}^{}_f(q^{}_\theta, \mathcal{D})  = \Ex_{q^{}_\theta(z)} \left [ f'\left( \frac{q^{}_\theta(z)}{p(z, \mathcal{D})}\right) \cdot \nabla^{}_{\theta} \log q^{}_\theta(z) \right],
	\end{equation}
	where $f'(t)$ denotes $\partial f(t) / \partial t$. Since $\nabla^{}_{\theta} \log q^{}_\theta(z)$ is known as the score function in statistics~\cite{Cox_1979} and is a part of the REINFORCE algorithm~\cite{Williams_ML_1992, Mnih_ICML_2014}, \eqref{score_fun_gradient} is called score function or REINFORCE gradient. An unbiased Monte Carlo (MC) estimator for~\eqref{score_fun_gradient} can be obtained by drawing $z_1, z_2, \cdots, z_K$ from $q_\theta^{}(z)$ and 
	\begin{equation}\label{score_fun_gradient_estimator}
		\nabla^{}_{\theta} \mathcal{\hat L}^{}_f(q^{}_\theta, \mathcal{D})  = \frac{1}{K} \sum_{k=1}^{K} \left[  f'\left( \frac{q^{}_\theta(z_k)}{p(z_k, \mathcal{D})}\right) \cdot \nabla^{}_{\theta} \log q^{}_\theta(z_k)  \right].
	\end{equation}
	However, since the variance of estimator~\eqref{score_fun_gradient_estimator} can be too large to be useful in practice, the score function gradient is usually employed along with some variation reduction techniques, such as the control variates and Rao-Blackwellization~\cite{Paisley_ICML_2012, Mnih_ICML_2014, Ranganath_ICAIS_2014}.

	An alternative to the score function gradient is the reparameterization gradient, which empirically has a lower estimation variance~\cite{Kingma_ICLR_2014, Zhang_arxiv_2019} and can be integrated with neural networks. The reparameterization trick requires the existence of a noise variable $\varepsilon\sim\allowbreak p(\varepsilon)$ and a mapping $g_\theta(\cdot)$ such that $z = g_\theta(\varepsilon)$. Instead of directly sampling $\{z_k\}_{k=1}^{K}$ from $q_\theta(z)$, the reparameterization estimators rely on the samples $\{\varepsilon_k\}_{k=1}^{K}$ drawn from $p(\varepsilon)$, for example, a Gaussian latent variable $z \sim q^{}_\theta(z) = \mathcal{N}(\mu, \Sigma)$ can be reparameterized with a standard Gaussian variable $\varepsilon \sim  \mathcal{N}(0, 1)$ and a mapping $z = g^{}_{\theta}(\varepsilon) = \mu + \Sigma^{\frac{1}{2}} \varepsilon$. More detailed interpretations as well as recent advances in the reparameterization trick can be found in~\cite{Kingma_ICLR_2014, Ruiz_NIPS_2016, Figurnov_NIPS_2018, Jankowiak_ICML_2018}. The gradient of $\mathcal{L}_f(q, \mathcal{D})$ after reparameterization becomes
	\begin{equation}\label{repar_fbound}
		\nabla_\theta \mathcal{L}^{\textrm{rep}}_f(q^{}_\theta, \mathcal{D}) = \nabla_\theta^{} \mathbb{E}_{p(\varepsilon)}\left[ f^*\left( \frac{p(g_\theta^{}(\varepsilon), \mathcal{D})}{q^{}_\theta(g_\theta^{}(\varepsilon))}  \right)  \right]. 
	\end{equation}
	An unbiased MC estimator for~\eqref{repar_fbound} is
	\begin{equation}\label{repar_estimator}
		\nabla_\theta^{} \hat{\mathcal{L}}^{\textrm{rep}}_{f}(q^{}_\theta, \mathcal{D}) =  \frac{1}{K} \sum_{k=1}^{K} \nabla_\theta^{}    f^*\left( \frac{p(g_\theta^{}(\varepsilon_k), \mathcal{D})}{q^{}_\theta(g_\theta^{}(\varepsilon_k))} \right) ,
	\end{equation}
	where $\varepsilon^{}_1, \varepsilon^{}_2, \cdots, \varepsilon^{}_K$ are drawn from $p(\varepsilon)$. We also give an unbiased MC estimator for the importance-weighted reparameterization gradient in~\eqref{IW_Estimator}, which will be utilized in later experiments: 
	\begin{equation}\label{IW_Estimator}
		\nabla_\theta^{} \hat{\mathcal{L}}^{\textrm{IW, rep}}_{f}(q^{}_\theta, \mathcal{D}, L) =  \frac{1}{K} \sum_{k=1}^{K} \nabla_\theta^{}    f^*\left( \frac{1}{L} \sum_{l=1}^{L} \frac{p(g_\theta^{}(\varepsilon_{k, l}), \mathcal{D})}{q^{}_\theta(g_\theta^{}(\varepsilon_{k, l}))} \right),
	\end{equation}
	where noise samples $\{\varepsilon_{k, 1:L}\}_{k=1}^{K}$ are drawn from $p(\varepsilon)$. All the aforementioned estimators for $f$-variational bounds and gradients are unbiased, while composing these estimator with other functions, \textit{e.g.} inverse dual functions in~\eqref{Sandwich_Bound}, can sometimes trade the unbiasedness for numerical stability~\cite{Li_NIPS_2016, Dieng_NIPS_2017, Tao_ICML_2018}.

	Nonetheless, the preceding estimators and VI algorithms rely on the full dataset $\mathcal{D}$ and can be handicapped to tackle the problems with large datasets. Meanwhile, since the properties of $f^*$-functions are flexible, it is non-trivial to represent the $f$-variational bounds by the expectation on a datapoint-wise loss, except for some specific divergences, such as KL divergence~\cite{Kingma_ICLR_2014} or divergences with dual functions $f^*$ satisfying $f^*(t \tilde{t}) = f^*(t) + f^*(\tilde{t})$, \textit{i.e.} $f^* \in \mathcal{F}_0$ with $\gamma = 0$. Therefore, to deploy the mini-batch training, we integrate the aforementioned estimators with the \textit{average likelihood} technique~\cite{Li_NIPS_2016}. Given a mini-batch of $M$ datapoints $\mathcal{D}_M = \{{x}_{n1}, \cdots, {x}_{nM}\} \subset \mathcal{D}$, we approximate the full log-likelihood by $\log p(\mathcal{D}|{z}) \approx N / M \cdot \sum_{m=1}^{M} \log p({x}_{nm}|{z})$. Hence, the ratio~$p(z, \mathcal{D}) / q(z)$ in~(\ref{score_fun_gradient}-\ref{IW_Estimator}) can be approximated by $\log [p({z}, \mathcal{D}) / q({z})] \approx N / M \cdot \sum_{m=1}^{M} \log p({x}_{nm}|{z}) + \log p(z) - \log q(z)$. When $z$ contains local hidden variables, the prior distribution $p(z)$ and approximate distribution $q(z)$ should also be approximated accordingly. This proxy to the full dataset wraps up our black-box $f$-VI algorithm, which is essentially a stochastic optimization algorithm that only relies on a mini-batch of data in each iteration. A reference black-box $f$-VI algorithm and the optimization schemes for a few concrete divergences are given in the SM.

	\subsection{Mean-field approximation}\label{sec3_3}
	Mean-field approximation, which simplifies the original VI problem for tractable computation, is historically an important VI algorithm before the emergence of stochastic VI. As the cornerstone of several variational message passing algorithms~\cite{Winn_JMLR_2005, Wand_BA_2011}, mean-field VI is still evolving~\cite{Knowles_NIPS_2011, Wang_JMLR_2013, Blei_JASA_2017, Zhang_TPAMI_2019} and worthy to be generalized for $f$-VI. A mean-field approximation assumes that all latent variables $\{z_j\}^{J}_{j=1}$ are independent, and the recognition model can be fully factorized as $q(z) = \prod_{j=1}^{J} q^{}_j(z_j)$, which simplifies the derivations and computation but might lead to less accurate results. The mean-field $f$-VI algorithm alternatively updates each marginal distribution $q_j$ to minimize the $f$-variational bound $\mathcal{L}_f(q, \mathcal{D})$. For the $f$-divergences with $f \in \mathcal{F}_1$, such as KL divergence, the coordinate-wise update rule for $q_j^{}(z_j)$ is obtained from fixing the other variational factors $q_{-j}^{}(z^{}_{-j}) = \prod_{\ell \neq j}q_\ell^{}(z^{}_\ell)$ and singling out $q_j(z_j)$ from $f$-variational bound $\mathcal{L}_f(q, \mathcal{D})$ in~\eqref{f_bound}, which gives
	\begin{equation}\label{rule1}
		q^*_j(z_j^{})\propto {f^*}^{-1}  \left( \Ex_{q^{}_{-j}} \left[ f^* \left( \frac{p(z, \mathcal{D})}{q^{}_{-j}(z_{-j}^{})} \right) \right] \right).
	\end{equation}
	For the $f$-divergences with $f \in \mathcal{F}_0$, such as $\chi$- or R\'enyi's $\alpha$-divergences, the coordinate-wise update rule for $q_j^{}(z_j)$ is obtained by applying the same procedures to the $f$-variational bound $\mathcal{L}_f(q, \mathcal{D}) = \mathbb{E}_{q(z)}[f(p(z, \mathcal{D}) / q(z))]$ from the forward $f$-VI (see SM), which gives
	\begin{equation}\label{rule2}
		q^*_j(z_j^{}) \propto  f^{-1}\left( \mathbb{E}_{q^{}_{-j}}\left[  f\left(\frac{p(z, \mathcal{D})}{q^{}_{-j}(z_{-j}^{})}\right)  \right] \right).
	\end{equation}
	When deriving these mean-field $f$-VI update rules (see SM), we only exploit the homogeneity of $f$- or $f^*$-function. CAVI~\cite{Bishop_2006, Blei_JASA_2017}, EP~\cite{Minka_UAI_2001}, and other types of mean-field VI algorithms can be restored from~\eqref{rule1} and \eqref{rule2} by choosing a proper $f$- or $f^*$-function. A reference mean-field VI algorithm along with a concrete realization example under KL divergence is provided in the SM. When the inverse function $f^{* -1}$ or $f^{-1}$ in~\eqref{rule1} or \eqref{rule2} is not analytically solvable, we can either generate a lookup table for $f^{* -1}$ or $f^{-1}$ and numerically evaluate \eqref{rule1} or \eqref{rule2} or resort to the stochastic $f$-VI.

	\section{Experiments}
	The effectiveness and the wide applicability of $f$-VI are demonstrated with three empirical examples in this section. We first verify the theoretical results with a synthetic example. The $f$-VI is then respectively implemented for a Bayesian neural network for linear regression and a VAE for image reconstruction and generation. Adam optimizer with recommended parameters in~\cite{Kingma_ICLR_2015} is employed for stochastic optimization, if not specified. Empirical results and data are reported by their mean value and $95\%$ confidence intervals. More detailed descriptions on the experimental settings, supplemental results, and the demonstration of the mean-field approximation method are provided in the SM.

	\subsection{Synthetic example}
\begin{figure}[htpb]
		\centering\vspace{-0.5em}
		\includegraphics[width=0.9\textwidth]{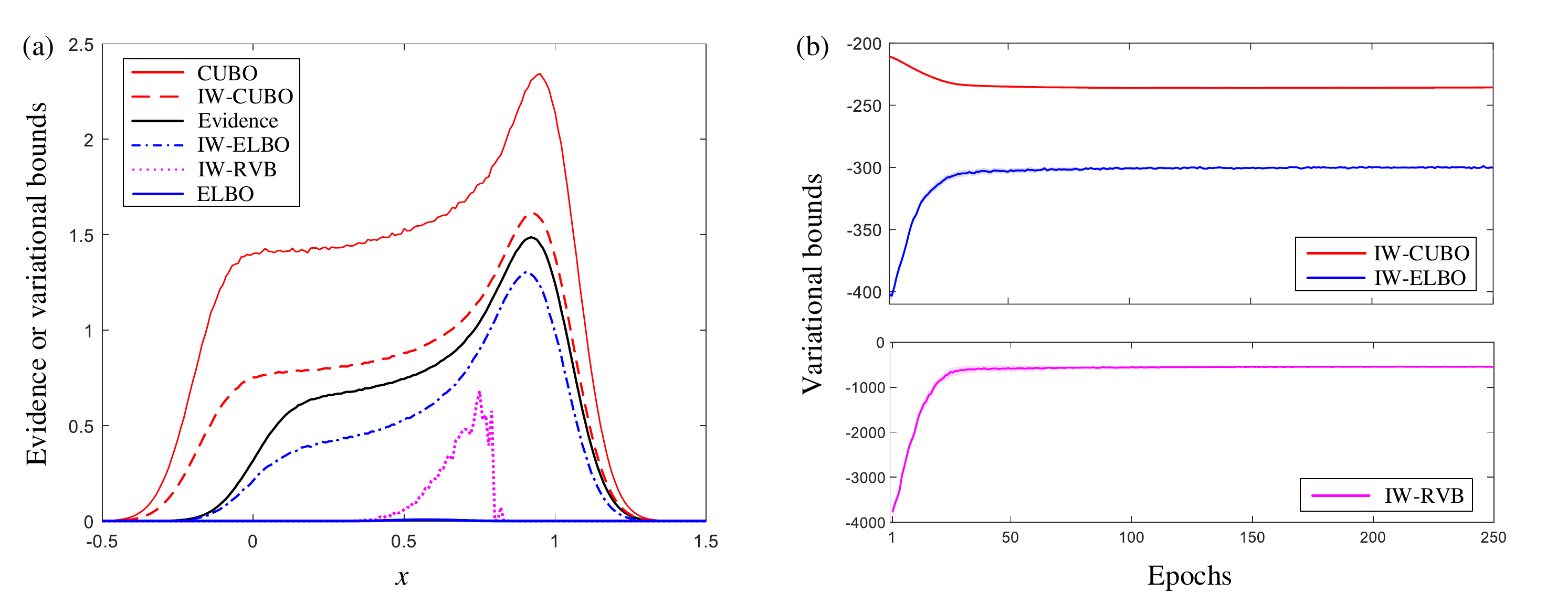}
		\caption{$f$-variational bounds on synthetic data.}\label{fig1}\vspace{-1.5em}
	\end{figure}

	We first demonstrate the $f$-VI theory with a vanilla example. Consider a batch of i.i.d. datapoints generated by a latent variable model $x = \sin(z) + \mathcal{N}(0, 0.01)$, $z \sim \textrm{UNIF}(0, \pi)$, where $\mathcal{N}(\mu, \sigma^2)$ denotes a univariate normal distribution with mean $\mu$ and variance $\sigma^2$, and $\textrm{UNIF}(a, b)$ denotes a uniform distribution on interval $[a, b]$. Subsequently, for simplicity, we posit a prior distribution $p(z) = \textrm{UNIF}(0, \pi)$, a likelihood distribution $p(x|z) = \mathcal{N}(\textrm{sin}(z), 0.01)$, and an approximate model $q^{}_\theta(z) = \textrm{UNIF}(\frac{1-\theta}{2} \pi, \frac{\theta + 1}{2} \pi)$, which is a uniform distribution centered at $z = \pi / 2$ with width $\theta \pi$. To verify the rank and the sharpness of $f$-variational bounds, we fix $\theta = 1.1$ and approximate the true evidence $p(x)$, IW-RVB ($\alpha = 2$), (IW-)CUBO ($n = 2$), and (IW-)ELBO ($L = 8$) in \hyperref[fig1]{Figure~1(a)}, which substantiates \hyperref[thm1]{Theorem~1},  \hyperref[cor0]{Corollary~1} and~\hyperref[cor1]{2}. A variational bound associated with the total variation distance, an $f$-divergence with non-monotonic $f^*$ function, is analyzed in the SM, and more approximation results when $q(z) = \mathcal{N}(\pi / 2, 1)$ can be found in~\cite{Tao_ICML_2018}. To demonstrate the effectiveness of stochastic $f$-VI algorithm, we set an initial value $\theta_0 = 1.5$ and update the recognition distribution $q_\theta^{}(z)$ by optimizing the IW-RVB ($\alpha = 3$), (IW-)CUBO ($n = 2$), and (IW-)ELBO. The IW-reparameterization gradient~\eqref{IW_Estimator} with $L = 3$ and $K = 1000$ is adopted for the training on a dataset of $500$ observations, and the $f$-variational bounds in \hyperref[fig1]{Figure~1(b)} are evaluated on a test set of $50$ observations. The sandwich-type bounds in \hyperref[fig1]{Figure~1(b)} give an estimate of the test log-evidence, which is roughly between $-235$ and $-300$.

	\subsection{Bayesian neural network}\label{sec4_2}
		We then implement the $f$-VI for a single-layer neural network for Bayesian linear regression. Our experimental setup generally follows the regression settings in~\cite{Li_NIPS_2016}, while some parameters vary to adapt to the $f$-VI framework. The linear regression is performed with twelve datasets from the UCI Machine Learning Repository~\cite{UCI}. Each dataset is randomly split into $90\% / 10\%$ for training and testing, and six different dual functions $f^*(\cdot)$ in $\mathcal{L}^{\text{\rm IW}}_{f}(q, \mathcal{D}, L)$ are selected such that three well-established $f$-VIs (KL-VI, R\'enyi's $\alpha$-VI with $\alpha = 3$, and $\chi$-VI with $n = 2$) and three new $f$-VIs (VIs subject to total variation distance and two custom $f$-divergences) are tested and compared. One of the custom $f$-divergences, inspired by~\cite{Bamler_NIPS_2017}, is defined by a convex dual function $f_{\textrm{c1}}^*(t) = \tilde{f}^*(t) - \tilde{f}^*(1)$, where $\tilde{f}^*(t) = -1/6 \cdot (\log t +t_0)^3 - 1/2 \cdot (\log t + t_0)^2 - (\log t + t_0) - 1$, $t = p(z, \mathcal{D}) / q(z)$, and $t_0 \in \mathbb{R}$ is a parameter to be optimized. The IW-reparameterization gradient with $L = 5$, $K = 50$ and mini-batch size of 32 is employed for training. After 20 trials with 500 training epochs in each trial, the regression results are evaluated by the test root mean squared error (RMSE) and test negative log-likelihood reported in~\hyperref[tab2]{Table~2}. The performance of custom $f_{\textrm{c1}}$-VI matches the results of well-established $f$-VIs on most datasets, and the custom $f_{\textrm{c1}}$-VI quantitatively outperforms others on some datasets, \textit{e.g.} Fish Toxicity and Stock. A complete version of \hyperref[tab2]{Table~2}, including the regression results of the other two new $f$-VIs, and more detailed descriptions on the training process, such as the architecture of neural network, training parameters, numerical stability and estimator biasedness, are provided in the SM. \vspace{-1em}	
		\setlength{\tabcolsep}{1.5pt}
		\begin{table}[H]\label{tab2}
			\newcommand\ColWidth{40}
			\caption{Average test error and negative log likelihood.}\vspace{0.2em}
			\centering	\footnotesize
			\begin{tabular}{ L{49 pt} C{\ColWidth pt} C{\ColWidth pt} C{\ColWidth pt} C{\ColWidth pt} C{\ColWidth pt} C{\ColWidth pt} C{\ColWidth pt} C{\ColWidth pt}}
				\toprule[1pt]
				\multirow{2}{*}{Dataset}	& \multicolumn{4}{c}{Test RMSE (lower is better)} & \multicolumn{4}{c}{Test negative log-likelihood (lower is better)}\\
				\cmidrule(lr){2-5}  \cmidrule(lr){6-9}
				& KL-VI & $\chi$-VI  & $\alpha$-VI & $f_{\textrm{c1}}$-VI &  KL-VI & $\chi$-VI  & $\alpha$-VI & $f_{\textrm{c1}}$-VI  \\   
				\cmidrule{1-9}  
				Airfoil & \bf{2.16$\pm$.07} & 2.36$\pm$.14 & 2.30$\pm$.08 & 2.34$\pm$.09 & \bf{2.17$\pm$.03} & 2.27$\pm$.03 & 2.26$\pm$.02 & 2.29$\pm$.02\\
				Aquatic & \bf{1.12$\pm$.06} & 1.20$\pm$.06 & 1.14$\pm$.07 & 1.14$\pm$.06 & \bf{1.54$\pm$.04} & 1.60$\pm$.08 & 1.54$\pm$.07 & 1.54$\pm$.06\\
				Boston &\bf{2.76$\pm$.36}& 2.99$\pm$.37 & 2.86$\pm$.36 &2.87$\pm$.36 & 2.49$\pm$.08 & 2.54$\pm$.18 & \bf{2.48$\pm$.13}  & 2.49$\pm$.13\\
				Building & 1.38$\pm$.12 & 2.82$\pm$.51 & 1.83$\pm$.22 & 1.80$\pm$.21 & 6.62$\pm$.02 & 6.94$\pm$.13 & 6.79$\pm$.03 & 6.74$\pm$.04\\
				CCPP & \bf{4.05$\pm$.09} & 4.14$\pm$.11 & 4.06$\pm$.08 & 4.33$\pm$.12 & \bf{2.82$\pm$.02} & 2.84$\pm$.03 & 2.82$\pm$.02 & 2.95$\pm$.01\\
				Concrete &5.40$\pm$.24& \bf{3.32$\pm$.34} & 5.32$\pm$.27 & 5.26$\pm$.21 & 3.10$\pm$.04 & \bf{2.61$\pm$.18} & 3.09$\pm$.04  & 3.09$\pm$.03\\
				Fish Toxicity & 0.88$\pm$.04 & 0.90$\pm$.04 & 0.89$\pm$.04 & 0.88$\pm$.03 & 1.28$\pm$.04 & 1.27$\pm$.04 & 1.29$\pm$.04 & 1.29$\pm$.03\\
				Protein & 1.93$\pm$.19 & 2.45$\pm$.42 & \bf{1.87$\pm$.17} & 1.97$\pm$.21 & \bf{2.00$\pm$.07} & 2.01$\pm$.08 & 2.04$\pm$.08 & 2.21$\pm$.04\\
				Real Estate &  7.48{$\pm$}1.41 & 7.51{$\pm$}1.44 & \bf{7.46{$\pm$}1.42} & 7.52{\scriptsize$\pm$}1.40 & 3.60{$\pm$}.30 & 3.70{$\pm$}.45 & \bf{3.59{$\pm$}.32} & 3.62{$\pm$}.33\\
				Stock & 3.85$\pm$1.12 & 3.90$\pm$1.09 & 3.88$\pm$1.13 & \bf{3.82$\pm$1.11} & -1.09$\pm$.04 & -1.09$\pm$.04 & -1.09$\pm$.04 & -1.09$\pm$.04\\
				Wine & .642$\pm$.018 & .640$\pm$.021 & .638$\pm$.018 & .643$\pm$.019 & .966$\pm$.027 & .965$\pm$.028 & .964$\pm$.025 & .975$\pm$.027\\
				Yacht & \bf{0.78$\pm$.12} & 1.18$\pm$.18 & 0.99$\pm$.12 & 1.00$\pm$.18 & \bf{1.70$\pm$.02} & 1.79$\pm$.03 & 1.82$\pm$.01 & 2.05$\pm$.01\\
				\bottomrule[1pt]
			\end{tabular}
		\end{table}

		\subsection{Bayesian variational autoencoder}\label{sec4_3}
		We also integrate the $f$-VI with a Bayesian VAE for image reconstruction and generation on the datasets of Caltech 101 Silhouettes~\cite{Caltech101}, Frey Face~\cite{FreyFace}, MNIST~\cite{MNIST}, and Omniglot~\cite{OMNIGLOT}. By replacing the conventional ELBO loss function of VAE~\cite{Kingma_ICLR_2014, MATLAB_VAE} with the more flexible $f$-variational bound loss functions, we test and compare the $f$-VAEs associated with three well-known $f$-divergences (KL-divergence, R\'enyi's $\alpha$-divergence with $\alpha = 3$, and $\chi$-divergence with $n = 2$) and three new $f$-divergences (total variation distance and two custom $f$-divergences). The dual function for total variation distance is $f^*(t) = |t-1|$. The custom $f_{\textrm{c1}}$-variational bound loss is induced by the aforementioned dual function $f_{\textrm{c1}}^*(t) = \tilde{f}^*(t) - \tilde{f}^*(1)$ with $t_0 = 0$. The custom $f_{\textrm{c2}}$-variational bound loss is induced by dual function $f_{\textrm{c2}}^*(t) = \log^2 t+ \log t $, which is convex on $t = p(z, \mathcal{D}) / q(z) \in(0, 1)$. The reparameterization gradient with $K =  3$, $L = 1$ is used for training. After 20 trials with 200 training epochs in each trial, the average test reconstruction errors (lower is better) measured by cross-entropy are listed in \hyperref[tab3]{Table~3}. In $f$-VAE example, the performances of three new $f$-VIs also rival the results of three well-known $f$-VIs on most datasets. Reconstructed and generated images, architectures of the encoder and decoder networks, and more detailed interpretations on the custom $f$-functions and training process of $f$-VAEs are given in the SM.
		\vspace{-1em}
		\setlength{\tabcolsep}{5.5pt}
		\begin{table}[H]
			\caption{Average test reconstruction errors of $f$-VAEs.}\vspace{0.2em}
			\label{tab3}
			\centering\footnotesize
			\begin{tabular}{lcccccc}
				\toprule[1pt]
				& KL-VI & $\chi$-VI & $\alpha$-VI & TV-VI & $f_{\textrm{c1}}$-VI & $f_{\textrm{c2}}$-VI\\
				\midrule[0.3pt]
				Caltech 101  & \bf{73.80}$\pm$2.27 & 73.84$\pm$2.16 & 74.95$\pm$2.76 &  74.32$\pm$2.26 & 74.87$\pm$2.56 &  74.85$\pm$2.94 \\
				Frey Face & 160.85$\pm$.72 & 160.57$\pm$.95 & 161.06$\pm$1.16 & 161.11$\pm$1.00 & \bf{160.52$\pm$.88} &  160.65$\pm$.87 \\
				MNIST & \bf{59.06}$\pm$.40 &  62.13$\pm$.50 & 61.90$\pm$.69 & 62.44$\pm$.41 & 59.60$\pm$.25 & 59.53$\pm$.42 \\
				Omniglot &    109.62$\pm$.20 &   110.57$\pm$.28 &   110.81$\pm$.32 & 110.21$\pm$.31 &  \bf{107.13$\pm$.39} & 108.29$\pm$.28    \\
				\bottomrule[1pt]
			\end{tabular}
		\end{table}

		\section{Conclusion}
		We have introduced a general $f$-divergence VI framework equipped with a rigorous theoretical analysis and a standardized optimization solution, which together extend the current VI methods to a broader range of statistical divergences. Empirical experiments on the popular benchmarks imply that this $f$-VI method is flexible, effective, and widely applicable, and some custom $f$-VI instances can attain state-of-the art results. Future work on $f$-VI may include finding the $f$-VI instances with more favorable properties, more efficient $f$-VI optimization methods, and  VI frameworks and theories that are more universal than the $f$-VI.
		
		\section*{Broader Impact}
		This work does not present any foreseeable societal consequence.

		\begin{ack}
			This work was supported by AFSOR under Grant FA9550-15-1-0518 and NSF NRI under Grant ECCS-1830639. The authors would like to thank the anonymous editors and reviewers for their constructive comments, Dr.~Xinyue Chang (Iowa State Univ.), Lei Ding (Univ.~of Alberta), Zhaobin Kuang (Stanford), Yang Wang (Univ.~of Alabama), and Yanbo Xu (Georgia Tech.) for their helpful suggestions, and Prof.~Evangelos A. Theodorou for his heuristic and insightful comments on this paper. In this arXiv version, the authors would also like to thank the readers and staff on arXiv.org.
		\end{ack}

	\small
	\setlength{\bibsep}{1pt}
	\bibliographystyle{unsrt}
	\bibliography{myref}
	
	\clearpage
	
	\normalsize
	
	\title{Supplementary Material for \\``$f$-Divergence Variational Inference''}

	\author{
		Neng Wan$^{\hspace{1pt}1\hspace{1pt}*}$\thanks{\ }\\
		\texttt{nengwan2@illinois.edu} \\
		\And
		Dapeng Li$^{\hspace{1pt}2\hspace{1pt}*}$\\
		\texttt{dapeng.ustc@gmail.com} \\
		\And
		Naira Hovakimyan$^1$ \\
		\texttt{nhovakim@illinois.edu} \\
	}
	
	\blfootnote{$^{*}$ Authors contributed equally to this paper.}


	\settitle
	\thispagestyle{empty}
	\vspace{-31pt}
	\begin{center}
		$^{1}$ University of Illinois at Urbana-Champaign, Urbana, IL 61801 \\
		$^{2}$ Anker Innovations, Shenzhen, China\\
	\end{center}
	\vspace{31pt}
	This supplementary material provides additional details for some results in the original paper.  
	
	\setcounter{section}{0}
	\renewcommand{\thesection}{\Alph{section}}
	\renewcommand*{\theHsection}{\Alph{section}}
	\section{Proofs of the main results}
	This section provides i) elaboration on the surrogate $f$-divergence including the proofs of \hyperref[prop1]{Proposition~1}, \hyperref[prop2]{Proposition~2} and \hyperref[prop3]{Proposition~3}, ii) deviations of the $f$-variational bound generated from both the reverse and forward surrogate $f$-divergence, and iii) an importance-weighted $f$-variational bound and the proof of \hyperref[cor0]{Corollary~1}.

	\subsection{Proof of Proposition 1}
	We first expand the LHS of~\eqref{prop1_eq1} by substituting the definitions of $f$-divergence~\eqref{F_Div} and generator function~\eqref{Generator_Fun}.
	\begin{equation*}
		\begin{split}
			\lim_{n \rightarrow \infty} D_{f_{\lambda_n^{}}}(q \| p) &  = \lim_{n \rightarrow \infty} \int p(z) \cdot \left[  f \left(\lambda_n \cdot \frac{q(z)}{p(z)}\right) - f(\lambda_n) \right]  \ dz \\
			& = \lim_{n \rightarrow \infty} \int p(z) \cdot f \left(\lambda_n \cdot \frac{q(z)}{p(z)}\right) \ dz  - \lim_{n \rightarrow \infty}  f(\lambda_n) \cdot \int p(z)    \ dz  \\
			& = \lim_{n \rightarrow \infty} \int p(z) \cdot f \left(\lambda_n \cdot \frac{q(z)}{p(z)}\right) \ dz. \\
		\end{split}
	\end{equation*}
	In order to prove~\eqref{prop1_eq1}, we only need to show that
	\begin{equation}\label{prop1_pf_eq1}
		\lim_{n \rightarrow \infty} \int p(z) \cdot f \left(\lambda_n \cdot \frac{q(z)}{p(z)}\right) \ dz = \int \lim_{n \rightarrow \infty} p(z) \cdot f \left(\lambda_n \cdot \frac{q(z)}{p(z)}\right) \ dz = D_f(q \| p),
	\end{equation}
	which can be proved by showing that function $g(\lambda) = \int p(x) \cdot f \left(\lambda \cdot q(z) / p(z)\right) dz$ is continuous in $\lambda$, since the continuity of $g(\lambda)$ brings each convergent sequence in $\lambda$ to a convergent sequence in $g(\cdot)$. The continuity of $g(\lambda)$ can be justified as follows. For arbitrary $\varepsilon>0$ and $z$, there exists $\delta  $ such that 
	\begin{equation*}
		\begin{split}
			\left| g(\lambda + \delta ) - g(\lambda)  \right| & = \left|   \int p(z) \cdot \left[ f\left( (\lambda +\delta) \cdot \frac{q(z)}{p(z)}  \right) - f\left( \lambda  \cdot \frac{q(z)}{p(z)}   \right)  \right]  dz  \right| \\
			& \leq   \int p(z) \cdot \left | f\left( (\lambda +\delta ) \cdot \frac{q(z)}{p(z)}  \right) - f\left( \lambda  \cdot \frac{q(z)}{p(z)}   \right)  \right | dz \\
			&\leq   \int p(z) \cdot  \epsilon \ dz=  \varepsilon\,,
		\end{split}
	\end{equation*}
	where we have used the uniform continuity of $f(\cdot)$. This completes the proof. \hfill\ensuremath{\blacksquare}

	\subsection{Proof of Proposition 2}
	We first consider the scenario when $f \in \mathcal{F}_0$. Since
	\begin{equation*}
		\begin{split}
			f^*(t \tilde{t}) & = t \tilde{t} \cdot f \left(\frac{1}{t\tilde{t}}\right)\\
			& = {t} \tilde t \cdot \left[ \left( \frac{1}{t} \right)^{\gamma_0} \cdot f\left( \frac{1}{\tilde t} \right) + f\left(\frac{1}{t}\right)  \right] \\
			& = t^{1 - \gamma_0} \cdot  f_0^*(\tilde t) + f_0^*(t) \cdot \tilde{t},
		\end{split}
	\end{equation*}
	by letting $\gamma = 1 - \gamma_0$, we can conclude that $f_0^* \in \mathcal{F}_1$. We then consider the case when $f \in \mathcal{F}_1$. Since
	\begin{align*}
		f^*(t\tilde{t}) & = t\tilde{t} \cdot f\left( \frac{1}{t\tilde{t}} \right)\\ \allowdisplaybreaks
		& = t\tilde{t} \cdot \left[ \left(\frac{1}{t} \right)^{\gamma_1} \cdot f\left(\frac{1}{\tilde t}\right) + f\left( \frac{1}{t} \right) \cdot \frac{1}{\tilde t}  \right]\\ \allowdisplaybreaks
		& = t^{1-\gamma_1} \cdot f_1^*(\tilde t) + f_1^*(t),
	\end{align*}
	by letting $\gamma = 1 - \gamma_1$, we can conclude that $f_1^* \in \mathcal{F}_0$. This completes the proof. \hfill\ensuremath{\blacksquare}

	\subsection{Proof of Proposition 3}
	We start this proof by substituting \eqref{F_Div},~\eqref{Generator_Fun} and~\eqref{ShiftHomo} into the LHS of~\eqref{thm1eq}
	\begin{equation*}
		\begin{split}
			D_{f_\lambda}(q \parallel p)&= \mathbb{E}_{p}[f_\lambda(q/p)]\\
			& =  \Ex_p [f (\lambda q /p)] - f(\lambda) \\
			&= \lambda^\gamma  \Ex_p  [ f(q/p) ] + f(\lambda) \cdot \Ex_p[ (p/q)^\eta] -f(\lambda).
		\end{split}
	\end{equation*}
	Since $f(\lambda) \cdot \Ex_p[ (p/q)^\eta] = f(\lambda) \cdot \Ex_p[ (p/q)^0 ] = f(\lambda)$ when $f \in \mathcal{F}_0$, and $f(\lambda) \cdot \Ex_p[ (p/q)^\eta] = f(\lambda)  \cdot \int q(x) \ dx = f(\lambda)$ when $f \in \mathcal{F}_1$, we have 
	\begin{equation*}
		D_{f_\lambda}(q \| p) = \lambda^\gamma  D_f(q \| p).
	\end{equation*}
	This completes the proof. \hfill\ensuremath{\blacksquare}
	
	\subsection{$f$-variational bound from reverse divergence}
	We provide detailed steps for deriving~\eqref{VB1}, which is a preliminary step for \hyperref[thm1]{Theorem~1} and the $f$-variational bound induced by reverse surrogate $f$-divergence. A reverse surrogate $f$-divergence can be decomposed as	
	\begin{equation*}
		\begin{split}
			D_{f_{p(\mathcal{D})^{-1}}}\left (q(z) \| p(z|\mathcal{D})\right ) & = \int p(z|\mathcal{D}) \cdot f_{p(\mathcal{D})^{-1}} \left( \frac{q(z)}{p(z|\mathcal{D})} \right) dz  \\
			& = \int p(z|\mathcal{D}) \cdot \left[ f \left( \frac{q(z) \cdot p(\mathcal{D})}{p(z, \mathcal{D})} \cdot \frac{1}{p(\mathcal{D})} \right) - f\left(\frac{1}{p(\mathcal{D})}\right) \right] dz  \\
			& =\frac{1}{p(\mathcal{D})}\int \frac{p(z, \mathcal{D})}{q(z)} \cdot  f \left( \frac{q(z) }{p(z, \mathcal{D})} \right) \cdot q(z) \ dz - f\left(\frac{1}{p(\mathcal{D})}\right) \\
			& = \frac{1}{p(\mathcal{D})} \cdot \Ex_{q(z)} \left[ f^*\left ( \frac{p(z, \mathcal{D})}{q(z)}\right ) \right] - f\left ( \frac{1}{p(\mathcal{D})} \right ). \\
		\end{split}
	\end{equation*}
	
	\subsection{$f$-variational bound from forward divergence}\label{sec_A5}
	As we mentioned in~\hyperref[sec3_1]{Section 3.1}, the assumption on $p(\mathcal{D}) > 0$ or the existence of $p(\mathcal{D})^{-1}$ in~\eqref{VB1} can be circumvented by using the $f$-VI that minimizes the forward surrogate $f$-divergence $D_{f_{p(\mathcal{D})}}(p(z|\mathcal{D})\|q(z))$. Meanwhile, in~\hyperref[sec3_3]{Section 3.3}, the coordinate-wise update rule~\eqref{rule2} for $f\in\mathcal{F}_0$ is also based on the $f$-variational bound induced by $D_{f_{p(\mathcal{D})}}(p(z|\mathcal{D})\|q(z))$. The $f$-variational bound and a sandwich estimate of evidence from forward surrogate $f$-divergence are derived below. First, we notice that the forward surrogate $f$-divergence can be decomposed as follows
	\begin{align*}
		D_{f_{p(\mathcal{D})}}(p(z|\mathcal{D}) \| q(z)) & = \int q(z) \cdot f_{p(\mathcal{D})}\left(  \frac{p(z|\mathcal{D})}{q(z)} \right)dz \\
		& = \int q(z) \cdot \left[ f\left( \frac{p(z, \mathcal{D})}{q(z) \cdot p(\mathcal{D})} \cdot p(\mathcal{D}) \right)  - f\left( p(\mathcal{D}) \right) \right] dz \nonumber\\ 
		& = \mathbb{E}_{q(z)}\left[ f\left( \frac{p(z, \mathcal{D})}{q(z)}  \right) \right] - f(p(\mathcal{D})). \nonumber
	\end{align*}	
	By the non-negativity of $f$-divergence~\cite{Sason_TIT_2016}, \textit{i.e.} $D_{f_{p(\mathcal{D})}}(p(z|\mathcal{D}) \| q(z)) \geq 0$, the $f$-variational bound $\mathcal{L}_f(q, \mathcal{D})$ from forward divergence follows
	\begin{equation}\label{forward_fbound}
		\mathcal{L}_f(q, \mathcal{D}) = \mathbb{E}_{q(z)}\left[ f\left( \frac{p(z, \mathcal{D})}{q(z)}  \right) \right] \geq f(p(\mathcal{D})),
	\end{equation}
	where equality holds when $q(z) = p(z|\mathcal{D})$. Inequality~\eqref{forward_fbound} formulates the $f$-variational bound induced by forward divergence $D_{f_{p(\mathcal{D})}}(p(z|\mathcal{D}) \| q(z))$ and supplements~\hyperref[thm1]{Theorem~1}, which is based on the reverse $f$-divergence. Given convex functions $f$ and $g$ such that $f(1) = g(1) = 0$,  on an interval where $f$ is non-decreasing and $g$ is non-increasing, a sandwich estimate of evidence $p(\mathcal{D})$ is given as follows
	\begin{equation*}
		(g)^{-1}\circ \Ex_{q(z)} \left[ g \left ( \frac{p(z,\mathcal{D})}{q(z)}\right ) \right]  \leq p(\mathcal{D}) \leq (f)^{-1}\circ \Ex_{q(z)} \left[ f \left( \frac{p(z, \mathcal{D})}{q(z)} \right) \right],
	\end{equation*}
	which supplements the sandwich estimate in \hyperref[cor1]{Corollary~2} derived from the reverse $f$-divergence. Stochastic $f$-VI algorithms that minimize $\mathcal{L}_f(q, \mathcal{D})$ in~\eqref{forward_fbound} can be readily implied by imitating the steps in  \hyperref[sec3_2]{Section 3.2}, and the optimization of $\mathcal{L}_f(q, \mathcal{D})$ in~\eqref{forward_fbound} also does not require $f$ and $g$ be invertible. Moreover, the statistical differences between $f$-variational bounds~\eqref{f_bound} and~\eqref{forward_fbound} deserve further investigations.

	\subsection{Proof of Corollary 1}
	The proof of \hyperref[cor0]{Corollary~1} is derived from the proof of Theorem~1 in the importance-weighted autoencoders paper~\cite{Burda_ICLR_2016}, and we will prove \hyperref[cor0]{Corollary~1} by utilizing the convexity of $f^*$-function and Jensen's inequality. First, we need to show that $\mathcal{L}^{\text{\rm IW}}_{f}(q, \mathcal{D}, L) \geq f^*(p(\mathcal{D}))$ for $L \in \mathbb{N}^*$, which is a direct result of Jensen's inequality
	\begin{align*}
		\mathcal{L}^{\text{\rm IW}}_{f}(q, \mathcal{D}, L) & = \Ex_{z_{1:L} \sim q(z)} \left[   f^*\left(  \frac{1}{L} \sum_{l=1}^{L} \frac{p(z_l, \mathcal{D})}{q(z_l)} \right) \right] \\
		& \geq f^*\left( \Ex_{z_{1:L} \sim q(z)} \left[ \frac{1}{L} \sum_{l=1}^{L} \frac{p(z_l, \mathcal{D})}{q(z_l)} \right]  \right) = f^*(p(\mathcal{D})).
	\end{align*}
	Next, we are to prove the statement that $\mathcal{L}^{\text{\rm IW}}_{f}(q, \mathcal{D}, L_1) \geq \mathcal{L}^{\text{\rm IW}}_{f}(q, \mathcal{D}, L_2)$ for $L_1 \leq L_2$. Let $\mathcal{I} = \{i_1, \cdots, i_{L_1}\} \subset \{1, 2, \cdots, L_2\}$ with $|\mathcal{I}| = L_1$ be a uniformly distributed subset of distinct indices from $\{1, 2, \cdots, L_2\}$. Subsequently, we have the identity $\mathbb{E}_{\mathcal{I} = \{ i_1, \cdots, i_m \}}[(a_{i_1} + \cdots + a_{i_{L_1}}) / L_1]$, which together with Jensen's inequality gives	
	\begin{align*}
		\mathcal{L}^{\text{\rm IW}}_{f}(q, \mathcal{D}, L_2)  & = \Ex_{z_{1:L_2} \sim q(z)} \left[   f^*\left(  \frac{1}{L_2} \sum_{l=1}^{L_2} \frac{p(z_l, \mathcal{D})}{q(z_l)} \right) \right] \\ \allowdisplaybreaks
		& = \Ex_{z_{1:L_2} \sim q(z)} \left[   f^*\left(  \mathbb{E}_{I = \{i_1, \cdots, i_{L_1} \}} \left[  \frac{1}{L_1} \sum_{l=1}^{L_1} \frac{p(z_l, \mathcal{D})}{q(z_l)}  \right]  \right) \right] \\
		& \leq \Ex_{z_{1:L_2} \sim q(z)} \left[  \mathbb{E}_{I = \{i_1, \cdots, i_{L_1} \}}\left[   f^*\left(  \frac{1}{L_1} \sum_{l=1}^{L_1} \frac{p(z_l, \mathcal{D})}{q(z_l)} \right)   \right]  \right] \\
		& = \Ex_{z_{1:L_1} \sim q(z)} \left[   f^*\left(  \frac{1}{L_1} \sum_{l=1}^{L_1} \frac{p(z_l, \mathcal{D})}{q(z_l)} \right) \right] = \mathcal{L}^{\text{\rm IW}}_{f}(q, \mathcal{D}, L_1). 
	\end{align*}
	Lastly, we need to show that $f^*(p(\mathcal{D})) = \lim_{L\rightarrow \infty}\mathcal{L}^{\text{\rm IW}}_{f}(q, \mathcal{D}, L)$, when $p(z, \mathcal{D}) / q(z)$ is bounded. Let the random variable $R_L = \frac{1}{L} \sum_{l=1}^{L}p(z_l, \mathcal{D}) / q(z_l)$ be bounded. By the strong law of large numbers, $R_L$ converges to $\mathbb{E}_{q(z_l)}[p(z_l, \mathcal{D}) / q(z_l)] = p(\mathcal{D})$ almost surely. Therefore, $\mathcal{L}^{\text{\rm IW}}_{f}(q, \mathcal{D}, L) = \mathbb{E}[f^*(R_L)]$ converges to $f^*(p(\mathcal{D}))$ \textit{a.s.} as $L\rightarrow \infty$. This completes the proof. 
	
	\section{Examples of $f$-variational bounds}
	In this section, we provide some concrete examples of $f$-variational bounds by using the relationship between $f$-divergence and some specific divergences~\cite{Sason_TIT_2016, Sason_Entropy_2018}. Some well-known variational bounds, such as ELBO~\cite{Jordan_ML_1999}, RVB~\cite{Li_NIPS_2016} and CUBO~\cite{Dieng_NIPS_2017}, are restored from $f$-variational bound~\eqref{f_bound}
	\begin{equation*}
		\mathcal{L}_{f}(q, \mathcal{D}) = \Ex_{q(z)} \left[ f^*\left( \frac{p(z, \mathcal{D})}{q(z)} \right) \right] \geq f^*(p(\mathcal{D})),
	\end{equation*}
	and some new bounds that have rarely been investigated for VI are also introduced.

	\subsection{$f$-variational bounds under KL divergence}
	The most famous variational bound induced by KL divergence is the ELBO. To restore ELBO from \eqref{f_bound}, consider a convex function $f(t) =  t \cdot \log t$ with $f(1) = 0$. Hence, the dual function $f^*(t) = -\log t$ with $f^*(t) = 0$ is convex and decreasing. Substituting this $f^*$-function into~\eqref{f_bound}, we have
	\begin{equation}\label{AppB_eq1}
		\log p(\mathcal{D}) \geq \mathbb{E}_{q(z)}[ \log p(z, \mathcal{D})] - \mathbb{E}_{q(z)}[\log q(z)] = \textrm{ELBO},
	\end{equation}
	where the RHS terms are known as the ELBO~\cite{Blei_JASA_2017}. Composing both sides of~\eqref{AppB_eq1} with an exponential function, we have a lower bound of evidence 
	\begin{equation*}
		p(\mathcal{D}) \geq \exp\left(  \mathbb{E}_{q(z)}[\log p(z, \mathcal{D})] - \mathbb{E}_{q(z)}[\log q(z)]  \right),
	\end{equation*}
	which verify the observation \textit{o}2) and \hyperref[cor1]{Corollary~2}. 
	
\begin{figure}[htpb]
		\centering
		\vspace{-1em}\includegraphics[width=0.55\textwidth]{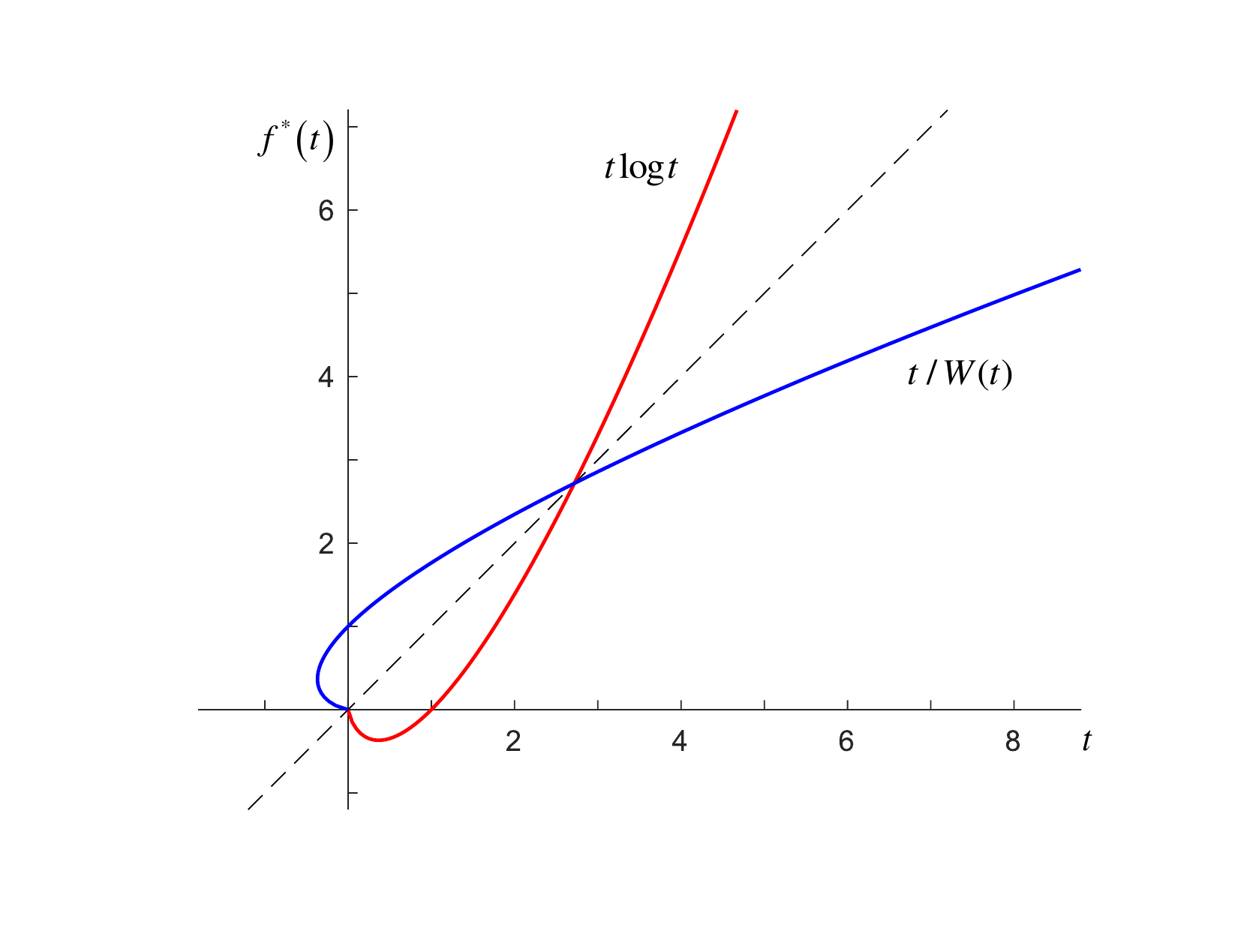}\vspace{-1.5em}
		\caption{$t \log t$ and its inverse function $t / W(t)$.}\label{fig_xlogx}
	\end{figure}

	While variational upper bounds of evidence have been already discovered in R\'enyi's $\alpha$-VI~\cite{Li_NIPS_2016} and $\chi$-VI~\cite{Dieng_NIPS_2017}, we rarely associate the variational upper bound with the classical KL-VI~\cite{Jordan_ML_1999}. With the new findings in~\hyperref[cor1]{Corollary~2}, we can readily define a variational upper bound subject to KL divergence. Consider the $f$-function, $f(t) = -\log t$ with $f(1) = 0$, associated with the forward KL divergence~\cite{Murphy_2012, Nowozin_NIPS_2016, Zhang_arxiv_2019} subject to the $f$-divergence in \hyperref[def1]{Definition~1}. The dual function then becomes $f^*(t) = t\log t$, which is decreasing on $(0, e^{-1}]$ and increasing on $(e^{-1}, \infty)$ as shown in~\hyperref[fig_xlogx]{Figure~2}. Hence, substituting $f^*(t) = t\log t$ into~\eqref{f_bound}, the $f$-variational bound under forward KL divergence is 
	\begin{equation}\label{AppB_eq3}
		\textrm{EUBO} = \Ex_{q(z)} \left[ \frac{p(z,\mathcal{D}))}{q(z)}\log \left( \frac{p(z,\mathcal{D})}{q(z)}\right)  \right] \geq p(\mathcal{D}) \cdot \log p(\mathcal{D}) = 	f^*(p(\mathcal{D})),
	\end{equation}
	where the LHS term is named as evidence upper bound (EUBO). Since $f^*(t) = t\log t$ is increasing on $(e^{-1}, \infty)$, EUBO in \eqref{AppB_eq3} provides an upper bound estimate of evidence when $p(\mathcal{D}) \geq e^{-1}$, which can be judged from the value of ELBO. When $p(\mathcal{D}) < e^{-1}$, one should resort to other divergences, \textit{e.g.} $\chi$-divergence and R\'enyi's $\alpha$-divergence, instead of KL divergence for an upper bound of evidence. To derive an upper bound on evidence, we will only consider the occasion when $p(\mathcal{D}) \geq e^{-1}$ hereafter. According to \hyperref[cor1]{Corollary~2}, an upper bound of $p(\mathcal{D})$ can be defined by composing both sides of \eqref{AppB_eq3} with the inverse function of $(f^*)^{-1}(t) = t / W(t)$, which is plotted in~\hyperref[fig_xlogx]{Figure~2}, and can be formulated as $(f^*)^{-1}(t) = t / W(t)$, which is well-defined on $t > 0$, and $W(t)$ is Lambert $W$ function implicitly defined by $t = W(t) \cdot \exp(W(t))$. 
	
	Hence, when $p(\mathcal{D}) \geq e^{-1}$, an upper bound induced by KL divergence can be formulated as follows
	\begin{equation}
		p(\mathcal{D}) \leq \max\{ \textrm{EUBO} / W(\textrm{EUBO}), e^{-1}    \},
	\end{equation}
	where EUBO is defined in \eqref{AppB_eq3}.

	\subsection{$f$-variational bounds under $\chi$-divergence}\label{secB2}
	We then associate the $f$-variational bound~\eqref{f_bound} with the $\chi$-divergence, which will restore the CUBO introduced in~\cite{Dieng_NIPS_2017}. The $\chi$-VI framework and CUBO introduced in~\cite{Dieng_NIPS_2017} are based on minimizing the forward $\chi^n$-divergence $D_{\chi^n}(p || q) = \mathbb{E}_{q(z)}[(p(z, x) / q(z))^n - 1]$ for $n\geq 1$, which is different from the reverse $\chi^n$-divergence given in~\hyperref[tab1]{Table~1}. While it may be more straightforward to restore CUBO from the $f$-VI based on forward divergence introduced in \hyperref[sec_A5]{Section A.5} or invoking~\hyperref[prop1]{Proposition~1} and~\hyperref[prop2]{Proposition~2} to convert the forward $\chi^n$-divergence to a reverse divergence, we will stick to the $f$-variational bound~\eqref{f_bound} and show that it is general enough to restore the CUBO with a properly chosen $f$-function. Consider an $f$-function $f(t) = t^{-1} - t$, which is convex on $t >0$ and satisfies $f(1) = 0$. The dual function then becomes $f^*(t) = t^2 - 1$, which is increasing on $t > 0$. Substituting the dual function into~\eqref{f_bound}, we have
	\begin{equation}\label{secB2_eq1}
		\mathbb{E}_{q(z)} \left[ \left( \frac{p(z, x)}{q(z)} \right)^2 - 1 \right] \geq p(x)^2 - 1.
	\end{equation}
	Canceling the constant terms in~\eqref{secB2_eq1} and taking the logarithm of both sides, CUBO$_2$ follows
	\begin{equation*}
		\textrm{CUBO}_2 = \frac{1}{2} \log \mathbb{E}_{q(z)} \left[ \left(  \frac{p(z, x)}{q(z)}  \right)^2 \right] \geq \log p(\mathcal{D}).
	\end{equation*}
	To restore the more general CUBO$_n$ for $n \in \mathbb{R} \backslash (0, 1) $, we consider the $f$-function $f(t) = t^{1-n} -t$, which is convex on $t \geq 0$ and satisfies $f(1) = 0$. The corresponding dual function is $f^*(t) = t^n - 1$, which is increasing on $t > 0$ when $n \geq 1$ and decreasing on $t > 0$ when $n \leq 0$. Substituting the dual function into~\eqref{f_bound}, we have
	\begin{equation}\label{secB2_eq2}
		\mathbb{E}_{q(z)} \left[ \left( \frac{p(z, x)}{q(z)} \right)^n - 1 \right] \geq p(x)^n - 1.
	\end{equation}
	Canceling the constant terms in~\eqref{secB2_eq2} and taking the logarithm of both sides, CUBO$_n$ follows
	\begin{equation}\label{CUBO_n}
		\textrm{CUBO}_n = \frac{1}{n} \log \mathbb{E}_{q(z)} \left[ \left(  \frac{p(z, x)}{q(z)} \right)^n \right]  \geq \log p(\mathcal{D}),
	\end{equation}
	which gives an evidence upper bound when $n \geq 1$ and a lower bound when $n \leq 0$. When $n \in (0, 1)$, a negative sign should be added such that a valid divergence is constructed~\cite{Li_NIPS_2016}. When $n < 1$, CUBO$_n$ recovers the RVB in~\cite{Li_NIPS_2016}, which will also be briefly discussed in \hyperref[secb3]{Section B.3}. The extension to $\chi^n$-VI under reverse $\chi^n$-divergence is left to interested readers.

	\subsection{$f$-variational bounds under R\'enyi's $\alpha$-divergence}\label{secb3}
	
	The R\'enyi's $\alpha$-divergence is defined as follows
	\begin{equation*}
		D_\alpha(p\|q) = \frac{1}{\alpha - 1} \log \int p(z, x)^{\alpha}q(z)^{1-\alpha}dz,
	\end{equation*}
	where $\alpha \in (0, 1)  \cup (1, +\infty)$. When $\alpha \in (-\infty, 0] \cup \{1\}$, $D_\alpha(p\|q)$ is not a valid divergence, and we will not consider this scenario, while interested readers can refer to~\cite{Li_NIPS_2016} for details. Rigorously, R\'enyi's $\alpha$-divergence is not an $f$-divergence; however, as shown in \hyperref[tab1]{Table~1}, a one-to-one correspondence can be established between the R\'enyi's $\alpha$-divergence and Hellinger $\alpha$-divergence, which is an $f$-divergence. We first show that $f$-variational bound~\eqref{f_bound} can restore the RVB when $\alpha > 1$~\cite{Li_NIPS_2016}. For $\alpha >1$, consider an $f$-function $f(t) =  t^\alpha - t$, which is convex on $t > 0$ and satisfies $f(1) = 0$. The dual function then becomes $f^*(t) = t^{1-\alpha} - 1$, which is decreasing on $t > 0$. Substituting this dual function into~\eqref{f_bound} and canceling the constant terms give
	\begin{equation}\label{AppB3_Eq1}
		\mathbb{E}_{q(z)}\left[ \left(\frac{p(z, \mathcal{D})}{q(z)}\right)^{1-\alpha} \right] \geq p(\mathcal{D})^{1-\alpha}.
	\end{equation}
	Since $f(t) = t^\alpha - t$ is not convex when $\alpha\in(0, 1)$, for this instance, we then consider the function $f(t) = -t^{\alpha}+t$, which is convex on $t > 0$ and satisfies $f(1) = 0$. The dual function is $f^*(t) = -t^{1-\alpha} + 1$, which is decreasing on $t > 0$ and $\alpha \in (0, 1)$. Substituting this dual function into~\eqref{f_bound} and canceling the constant terms give
	\begin{equation}\label{AppB3_Eq2} 
		\mathbb{E}_{q(z)}\left[\left(\frac{p(z, \mathcal{D})}{q(z)}\right)^{1-\alpha} \right] \leq p(\mathcal{D})^{1-\alpha}.
	\end{equation}
	Taking the logarithm on both sides of~\eqref{AppB3_Eq1} and~\eqref{AppB3_Eq2}, and dividing both sides of the results by $1-\alpha$, we have
	\begin{equation}\label{RVB}
		\textrm{RVB} = \frac{1}{1-\alpha} \log \mathbb{E}_{q(z)}\left[\left(\frac{p(z, \mathcal{D})}{q(z)}\right)^{1-\alpha} \right] \leq \log p(\mathcal{D}),
	\end{equation}
	which is identical to the RVB $\mathcal{L}_{\alpha+}(q; \mathcal{D})$ defined in~\cite{Li_NIPS_2016}.

	\subsection{$f$-variational bounds under total variation distance}
	The total variation distance is induced by the $f$-function $f(t) = |t-1|$ with dual function $f^*(t) = |t-1| = f(t)$. This $f$-function poses stark differences than the previous examples: i) $f$- and $f^*$-functions are not smooth at $t = 1$, ii) $f$- and $f^*$-functions are not monotonic on $t > 0$, and iii) the dual function $f^*(t) = |t-1|$ is not invertible. Nonetheless, since the dual function $f^*(t) = |t-1|$ is decreasing on $t\in[0, 1)$ and increasing on $t\in(1, \infty)$, the $f$-variational bounds subject to total variation can still provide a valid upper/lower bound of evidence on each monotonic interval. Substituting $f^*(t) = |t-1| = f(t)$ into~\eqref{f_bound}, we have
	\begin{equation}\label{secb4_eq1}
		\Ex_{q(z)} \left[ \left| \frac{p(z, \mathcal{D})}{q(z)} - 1 \right| \right]	\geq |p(\mathcal{D}) - 1|.
	\end{equation}
	When $p(\mathcal{D}) \in [0, 1)$, inequality~\eqref{secb4_eq1} gives a lower bound of evidence
	\begin{equation}\label{secb4_eq2}
		p(\mathcal{D}) \geq  1 - \Ex_{q(z)} \left[ \left| \frac{p(z, \mathcal{D})}{q(z)} - 1 \right| \right].
	\end{equation}
	When $p(\mathcal{D}) \geq 1$, inequality~\eqref{secb4_eq1} gives an upper bound of evidence
	\begin{equation}\label{secb4_eq3}
		p(\mathcal{D}) \leq  1 + \Ex_{q(z)} \left[ \left| \frac{p(z, \mathcal{D})}{q(z)} - 1 \right| \right].
	\end{equation}
	Combining~\eqref{secb4_eq2} and~\eqref{secb4_eq3}, the $f$-variational bounds induced by the total variation distance are given as follows
	\begin{equation}\label{TVB}
		\max\left\{0, 1 - \Ex_{q(z)} \left[ \left| \frac{p(z, \mathcal{D})}{q(z)} - 1 \right| \right] \right\} \leq p(\mathcal{D}) \leq 1 + \Ex_{q(z)} \left[ \left| \frac{p(z, \mathcal{D})}{q(z)} - 1 \right| \right].
	\end{equation}
	A vanilla example demonstrating the $f$-variational bounds associated with total variation distance is provided in~\hyperref[fig3]{Figure~3} of \hyperref[secE_1]{Section E.1}.

	\section{Examples of stochastic $f$-variational inference}
	This section provides supplementary interpretations for \hyperref[sec3_2]{Section~3.2} with i) steps for deriving the score function gradient in~\eqref{score_fun_gradient}, ii) concrete examples of the score function, reparameterization, and IW-reparameterization gradients under KL, $\chi$-, and R\'enyi's $\alpha$-divergences, and iii) a reference algorithm for black box (stochastic) $f$-VI. First, we derive the score function gradient \eqref{score_fun_gradient} for optimizing the parameters $\theta$ in recognition model $q_\theta(z)$. Computing the gradient of $f$-variational bound $\mathcal{L}_f(q_\theta, \mathcal{D})$ in~\eqref{f_bound} w.r.t. parameters $\theta$, we have
	\begin{equation*}
		\begin{split}
			\nabla^{}_{\theta} \mathcal{L}_f(q^{}_\theta, \mathcal{D})  = \nabla^{}_{\theta} \Ex_{q^{}_\theta(z)} \left[  f^*\left ( \frac{p(z, \mathcal{D})}{q^{}_\theta(z)}\right )  \right]   & =  \int p(z, \mathcal{D}) \cdot \nabla^{}_{\theta} f \left( \frac{q^{}_\theta(z)}{p(z, \mathcal{D})}\right) dz \\
			& = \int q^{}_\theta(z) \cdot f' \left( \frac{q^{}_\theta(z)}{p(z, \mathcal{D})}\right) \cdot \frac{\nabla^{}_{\theta} q^{}_\theta(z)}{q^{}_\theta(z)} \ dz  \\
			& = \Ex_{q^{}_\theta(z)} \left [ f'\left( \frac{q^{}_\theta(z)}{p(z, \mathcal{D})}\right) \cdot \nabla^{}_{\theta} \log q^{}_\theta(z) \right ] ,
		\end{split}
	\end{equation*}
	where $f'(t)$ denotes $\partial f(t) / \partial t$. An unbiased MC estimator for this score gradient function is given in~\eqref{score_fun_gradient_estimator}.

	\subsection{Gradient estimators under KL divergence}
	We first provide the gradient estimators for stochastic $f$-VI subject to KL divergence. For the ELBO originated from reverse KL divergence, we choose the $f$-function $f(t)= t\log t$, which gives the dual function $f^*(t) = -\log t$ and derivative $f'(t)=1+\log t$. Substituting $f'(t)=1+\log t$ into~\eqref{score_fun_gradient_estimator} and multiplying the result by $-1$\footnote{When deriving the ELBO in~\eqref{AppB_eq1}, we also multiplied the $f$-variational bound by $-1$.}, we have a score function gradient estimator of ELBO
	\begin{equation}\label{secC1_eq1}
		\nabla_\theta^{} \mathcal{\hat L}_f(q_\theta, \mathcal{D}) = \frac{1}{K} \sum_{k=1}^{K} \log\frac{p(z_k, \mathcal{D})}{q_\theta^{}(z_k)}  \cdot \nabla_\theta^{} \log q^{}_\theta(z_k),
	\end{equation}
	where $z_k \sim q_\theta^{}(z)$. The score function gradient estimator~\eqref{secC1_eq1} for ELBO restores the result in~\cite{Mnih_ICML_2014}. Given a noise variable $\varepsilon \sim p(\varepsilon)$ and a mapping $g_\theta(\cdot)$ such that $z = g_\theta(\varepsilon)$, and substituting $f^*(t) = -\log t$ into~\eqref{repar_estimator} and multiplying the result by $-1$, we have a reparameterization gradient estimator of ELBO
	\begin{equation}\label{secC1_eq2}
		\nabla_\theta^{}\mathcal{\hat L}_f^{\textrm{rep}}(q_\theta, \mathcal{D}) = \frac{1}{K} \sum_{k=1}^{K} \nabla_\theta^{} \log \frac{p(g^{}_\theta(\varepsilon_k), \mathcal{D})}{q(g^{}_\theta(\varepsilon_k)) } ,
	\end{equation}
	where $\varepsilon_k \sim p(\varepsilon)$. The reparameterization gradient~\eqref{secC1_eq2} restores the gradient of standard VAE in~\cite{Kingma_ICLR_2014}. Substituting $f^*(t) = -\log t$ into~\eqref{IW_Estimator} and drawing the two-dimensional noise samples $\{\varepsilon_{k, 1:L}\}_{k=1}^{K}$ from $p(\varepsilon)$, we have an IW-reparameterization gradient of ELBO
	\begin{equation*}
		\nabla_\theta^{} \hat{\mathcal{L}}^{\textrm{IW, rep}}_{f}(q^{}_\theta, \mathcal{D}, L) = \frac{1}{K} \sum_{k=1}^{K}  \nabla_\theta^{} \log \left( \frac{1}{L} \sum_{l=1}^{L} \frac{p(g_\theta^{}(\varepsilon_{k, l}), \mathcal{D})}{q^{}_\theta(g_\theta^{}(\varepsilon_{k, l}))} \right),
	\end{equation*}
	which restores the gradient of IW-VAE in~\cite{Burda_ICLR_2016}. In practice, the (IW-)reparameterization gradients can be computed by invoking the backpropagation functions in machine learning libraries or other automatic differentiation tools.

	We then give the gradients for optimizing the EUBO defined in~\eqref{AppB_eq3}, which has rarely been reported before. For EUBO, we consider the $f$-function $f(t) = -\log t$, which gives the dual function $f^*(t) = t \log t$ and derivative $f'(t) = -1 / t$. Hence, substituting $f'(t) = -1 / t$ into~\eqref{score_fun_gradient_estimator}, we have a score function gradient estimator of EUBO
	\begin{equation*}
		\nabla_\theta \mathcal{\hat L}_f(q^{}_\theta, \mathcal{D})  = - \frac{1}{K} \sum_{k = 1}^{K} \frac{p(z_k, \mathcal{D})}{q_\theta^{}(z_k)} \cdot \nabla_\theta^{} \log q_\theta^{}(z_k),
	\end{equation*}
	where $z_k \sim q_\theta^{}(z)$. The reparameterization gradient estimator of EUBO can be obtained by substituting the dual function $f^*(t) = t \log t$ into~\eqref{repar_estimator}, which gives
	\begin{equation*}
		\nabla_\theta \mathcal{\hat L}_{f}^{\textrm{rep}}(q_\theta^{}, \mathcal{D}) = \frac{1}{K} \sum_{k=1}^{K} \nabla_\theta^{} \left(  \frac{p(g_\theta^{}(\varepsilon_k),\mathcal{D}))}{q_\theta^{}(g_\theta^{}(\varepsilon_k))} \cdot \log \frac{p(g_\theta^{}(\varepsilon_k),\mathcal{D})}{q^{}_\theta(g_\theta^{}(\varepsilon_k))}       \right), 
	\end{equation*}
	where noise samples $\varepsilon_k \sim p(\varepsilon)$. The IW-reparameterization gradient estimator of EUBO is obtained by substituting the dual function $f^*(t) = t \log t$ into~\eqref{IW_Estimator}, which gives
	\begin{equation*}
		\nabla_\theta \mathcal{\hat L}_{f}^{\textrm{IW, rep}}(q_\theta^{}, \mathcal{D}, L) = \frac{1}{K} \sum_{k=1}^{K} \nabla_\theta^{} \left( \frac{1}{L} \sum_{l=1}^{L} \frac{p(g_\theta^{}(\varepsilon_{k, l}), \mathcal{D})}{q^{}_\theta(g_\theta^{}(\varepsilon_{k, l}))} \cdot \log \left(  \frac{1}{L} \sum_{l=1}^{L} \frac{p(g_\theta^{}(\varepsilon_{k, l}), \mathcal{D})}{q^{}_\theta(g_\theta^{}(\varepsilon_{k, l}))}  \right)       \right), 
	\end{equation*}
	where the noise samples $\{\varepsilon_{k, 1:L}\}_{k=1}^{K} \sim p(\varepsilon)$.

	\subsection{Gradient estimators under $\chi$-divergence}\label{secC2}
	We then implement the gradient estimators of $f$-VI to $\chi$-divergence. For conciseness, we only consider the gradient of objective function $\exp\left( n \cdot \textrm{CUBO}_n \right)$, which has unbiased estimators, while the estimators of $\textrm{CUBO}_n$ in~\eqref{CUBO_n} are biased but more stable in numerical computation. Similar to \hyperref[secB2]{Section B.2}, we choose the $f$-function $f(t) = t^{1-n}-t$, which implies the dual function $f^*(t) = t^n - 1$ and the derivative $f'(t) = (1-n) t^{-n} - 1$. Hence, substituting $f'(t) = (1-n) t^{-n} - 1$ into~\eqref{score_fun_gradient_estimator}, we have a score function gradient estimator for $\chi$-VI
	\begin{equation*}
		\nabla_\theta \mathcal{\hat L}_f(q_\theta^{}, \mathcal{D})  = \frac{1-n}{K}\sum_{k=1}^{K} \left[  \left( \frac{p(z_k, \mathcal{D})}{q^{}_\theta(z_k)} \right)^n \nabla_\theta \log q^{}_\theta(z_k) \right]
	\end{equation*}
	where $z_k \sim q_\theta^{}(z)$. Given a noise variable $\varepsilon \sim p(\varepsilon)$ and a mapping $g_\theta(\cdot)$ such that $z = g_\theta(\varepsilon)$, the reparameterization gradient estimator is obtained by substituting $f^*(t) = t^n - 1$ into~\eqref{repar_estimator}
	\begin{equation*}
		\nabla_\theta^{} \mathcal{\hat L}^{\textrm{rep}}_{f}(q^{}_\theta, \mathcal{D})  = \frac{1}{K} \sum_{k=1}^{K} \nabla_\theta \left(  \frac{p(g_\theta^{}(\varepsilon_k), x)}{q_\theta^{}(g_\theta^{}(\varepsilon_k))} \right)^n 
		= \frac{n}{K} \sum_{k=1}^{K}  \left(  \frac{p(g_\theta^{}(\varepsilon_k), x)}{q_\theta^{}(g_\theta^{}(\varepsilon_k))} \right)^n \nabla_\theta \log \frac{p(g_\theta^{}(\varepsilon_k), x)}{q_\theta^{}(g_\theta^{}(\varepsilon_k))} ,
	\end{equation*}
	where noise samples $\varepsilon_k \sim p(\varepsilon)$. While the preceding two gradient estimators recover the result in~\cite{Dieng_NIPS_2017}, we supplement $\chi$-VI with an IW-reparameterization gradient estimator, which is obtained by substituting $f^*(t) = t^n - 1$ into~\eqref{IW_Estimator}
	\begin{equation*}
		\nabla_\theta^{} \mathcal{\hat L}^{\textrm{IW, rep}}_{f}(q^{}_\theta, \mathcal{D}, L)  =  \frac{n}{K} \sum_{k=1}^{K}  \left(   \frac{1}{L} \sum_{l=1}^{L} \frac{p(g_\theta^{}(\varepsilon_{k, l}), \mathcal{D})}{q^{}_\theta(g_\theta^{}(\varepsilon_{k, l}))} \right)^n \nabla_\theta \log \left(   \frac{1}{L} \sum_{l=1}^{L} \frac{p(g_\theta^{}(\varepsilon_{k, l}), \mathcal{D})}{q^{}_\theta(g_\theta^{}(\varepsilon_{k, l}))}   \right) ,
	\end{equation*}
	where the noise samples $\{\varepsilon_{k, 1:L}\}_{k=1}^{K} \sim p(\varepsilon)$.

	\subsection{Gradient estimators under R\'enyi's $\alpha$-divergence}
	Our last example implements the $f$-VI gradient estimators to R\'enyi's $\alpha$-divergences and supplements R\'enyi's $\alpha$-VI~\cite{Li_NIPS_2016} with a set of unbiased gradient estimators. Similar to the gradients of $\chi$-VI introduced in \hyperref[secC2]{Section~C.2}, this section considers the gradient estimators of objective function $\exp\{(1-\alpha) \cdot \textrm{RVB}\}$, where RVB is defined in~\eqref{RVB}. The choices of $f$-functions are i) $f(t) = t^\alpha - t$, $f^*(t) = t^{1-\alpha} -1$ and $f'(t) = \alpha t^{\alpha - 1} - 1$  for $\alpha \in (0, 1)$, and ii) $f(t) = -t^\alpha + t$, $f^*(t) = - t^{1-\alpha} + 1$, and $f'(t) = -\alpha t^{\alpha - 1} + 1$ for $\alpha \in (1, +\infty)$. Consequently, the score gradient estimator is
	\begin{equation*}
		\nabla_\theta \mathcal{\hat L}_f(q_\theta^{}, \mathcal{D}) = \frac{\alpha}{K} \sum_{k=1}^{K} \left( \frac{q^{}_\theta(z_k)}{p(z_k, \mathcal{D})} \right)^{\alpha - 1} 	\nabla_\theta \log q^{}_\theta(z_k),
	\end{equation*}
	where $z_k \sim q_\theta^{}(z)$. Given a noise variable $\varepsilon \sim p(\varepsilon)$ and a mapping $g_\theta(\cdot)$ such that $z = g_\theta(\varepsilon)$, the reparameterization gradient estimator under R\'enyi's $\alpha$-divergence is
	\begin{equation*}
		\nabla_\theta \mathcal{\hat L}_f^{\textrm{rep}}(q_\theta^{}, \mathcal{D}) =  \frac{1-\alpha}{K} \sum_{k=1}^{K} \left(  \frac{p(g_\theta(\varepsilon_k), \mathcal{D})}{q^{}_\theta(g_\theta(\varepsilon_k))} \right)^{1-\alpha}   \nabla_\theta^{} \log \frac{p(g_\theta^{}(\varepsilon_k), x)}{q_\theta^{}(g_\theta^{}(\varepsilon_k))} ,
	\end{equation*}
	where noise samples $\varepsilon_k \sim p(\varepsilon)$. The IW-reparameterization gradient estimator then becomes 
	\begin{equation*}
		\nabla_\theta \mathcal{\hat L}_f^{\textrm{rep}}(q_\theta^{}, \mathcal{D}, L) =  \frac{1-\alpha}{K} \sum_{k=1}^{K} \left(  \frac{1}{L} \sum_{l=1}^{L} \frac{p(g_\theta^{}(\varepsilon_{k, l}), \mathcal{D})}{q^{}_\theta(g_\theta^{}(\varepsilon_{k, l}))} \right)^{1-\alpha}   \nabla_\theta^{} \log \left( \frac{1}{L} \sum_{l=1}^{L} \frac{p(g_\theta^{}(\varepsilon_{k, l}), \mathcal{D})}{q^{}_\theta(g_\theta^{}(\varepsilon_{k, l}))}  \right) ,
	\end{equation*}
	where the noise samples $\{\varepsilon_{k, 1:L}\}_{k=1}^{K} \sim p(\varepsilon)$.

	\subsection{Stochastic $f$-variational inference algorithm}
	The following table provides a reference algorithm to implement stochastic $f$-VI.
	
	\setlength{\algoheightrule}{0.75pt} 
	\setlength{\algotitleheightrule}{0.3pt} 
	\begin{algorithm}[H]
		\caption{Stochastic $f$-VI}
		\SetAlgoLined
		\KwIn{Dataset $\mathcal{D} = \{x_n\}_{n=1}^N$,	model $p(z, x)$, variational family $q^{}_{{\theta}}({z})$, and $f$-function.}
		\renewcommand{\KwIn}{\textbf{Initialize: }} 
		\KwIn{Recognition parameters $\theta_0$.}\\
		\While{$\theta$ has not converged}{
			Randomly draw i) a minibatch $\mathcal{D}_M$ from full dataset $\mathcal{D}$ and ii) nosise samples $\{\varepsilon_k\}_{k=1}^{K}$ or $\{\varepsilon_{k, 1:L}\}_{k=1}^{K}$ from noise distribution $p(\bm{\varepsilon})$\;
			Approximate full likelihood $p(g^{}_{\theta_t}(\varepsilon_k), D)$, recognition distribution $q^{}_{\theta_t}(g^{}_{\theta_t}(\varepsilon_k))$, and prior distribution $p(g^{}_{\theta_t}(\varepsilon_k))$\;
			Compute the gradient of $f$-variational bound from~\eqref{score_fun_gradient_estimator}, \eqref{repar_estimator} or \eqref{IW_Estimator}\;
			Update parameters $\theta_{t+1}$ from $\theta_t$ and the gradient.}
		\renewcommand{\KwResult}{\textbf{Return: }} 
		\KwResult{Recognition distribution $q^{}_\theta(z)$.}
	\end{algorithm}

	\section{Mean-field $f$-variational inference}
	
	This section supplements the mean-field $f$-VI by providing i) steps for deriving the coordinate-wise update rules~\eqref{rule1} and~\eqref{rule2}, ii) an example of mean-field $f$-VI subject to KL divergence, and iii) a reference mean-field $f$-VI algorithm. The mean-field $f$-VI is developed on the basis of mean-field assumption $q(z) = \prod_{j=1}^{J}q_j(z_j)$ and $f$-function's homogeneity decomposition $f \in \mathcal{F}_{\{0, 1\}}$.

	\subsection{Deviation of update rules}
	We first show the detailed steps for deriving the coordinate-wise update rules~\eqref{rule1} and~\eqref{rule2} in mean-field $f$-VI. For conciseness, we define $p = p(z, \mathcal{D})$ and $q = q(z) = q_j(z_j) \cdot \prod_{\ell\neq j} q_\ell(z_\ell) = q_j \cdot q_{-j}$. The update rules are then derived by singling out the term $q_j$ from $f$-variational bounds~\eqref{f_bound} or~\eqref{forward_fbound} while fixing all the other terms that consist of $q_{-j}$. For $f$-divergences with $f \in \mathcal{F}_1$ or $f^* \in \mathcal{F}_0$, such as KL divergence, we have $f(t\tilde{t}) = t^{\gamma} f(\tilde{t}) + f(t) \tilde{t}$ and $f^*(t\tilde{t}) = t^{1-\gamma} f^*(\tilde{t}) + f^*(t)$. Hence, the $f$-variational bound~\eqref{f_bound} can be reformulated as
	\begin{equation*}
		\begin{split}
			\mathcal L_{f}(q_j^{}, q_{-j}^{}, \mathcal{D}) & =  \Ex_q \left[ \frac{p}{q} \cdot f\left(\frac{q}{p}\right)  \right] = \mathbb{E}_q\left[f^*\left(  \frac{1}{q_j} \cdot \frac{p}{q_{-j}} \right)  \right] \\
			& = \Ex_q \left[ q_j^{\gamma -1} \cdot f^*\left( \frac{p}{q_{-j}} \right) \right]+ \mathbb{E}_q \left[ f^*\left(\frac{1}{q_j}\right) \right]    \\
			& = \mathbb{E}_{q_j}\left[q_j^{\gamma - 1} \cdot \mathbb{E}_{q_{-j}} \left[ f^*\left( \frac{p}{q_{-j}} \right) \right] \right] + \mathbb{E}_{q_j}\left[ f^*\left( \frac{1}{q_j} \right) \right] \\
			& = \mathbb{E}_{q_j}\left[q_j^{\gamma - 1} \cdot f^* \circ f^{*-1}\left( \mathbb{E}_{q_{-j}} \left[ f^*\left( \frac{p}{q_{-j}} \right) \right] \right) + \frac{1}{q_j} \cdot f(q_j) \right] \\
			& = \mathbb{E}_{q_j}\left[ \frac{m_j}{q_j} \cdot \left( q_j^\gamma \cdot  f\left(\frac{1}{m_j}\right) + f(q_j) \cdot \frac{1}{m_j}   \right)   \right] \\
			& = \mathbb{E}_{q_j}\left[ \frac{m_j}{q_j} \cdot f\left( \frac{q_j}{m_j} \right) \right] = \mathbb{E}_{q_j}\left[f^*\left( \frac{m_j}{q_j} \right) \right] ,
		\end{split}
	\end{equation*}
	where $m^{}_j = {f^*}^{-1}  ( \Ex_{q^{}_{-j}} [ f^* (p / q_{-j}^{})])$ can be regarded as an unnormalized probability distribution. After normalizing $m_j$ into a probability distribution $\tilde{m}_j$ with normalization constant $c > 0$, the $f$-variational bound then becomes $\mathcal L_{f}(q_j^{}, q_{-j}^{}, \mathcal{D})  = c \cdot D^{}_{f^*}(\tilde{m}_j \| q_j)$, which attains its minimum at $\tilde{m}_j = q_j$. Therefore, to minimize the $f$-variational bound when $f \in \mathcal{F}_1$, the marginal distribution $q_j$ should be updated in accordance with~\eqref{rule1}:
	\begin{equation*}
		q_j^{} \propto m_j = {f^*}^{-1}  \left( \Ex_{q^{}_{-j}} \left[ f^* \left( \frac{p(z, \mathcal{D})}{q^{}_{-j}(z_{-j}^{})} \right) \right] \right).
	\end{equation*}

	For $f$-divergences with $f\in\mathcal{F}_0$ or $f^* \in \mathcal{F}_1$, such as $\chi$- and R\'enyi's $\alpha$-divergences, we have identity $f(t\tilde{t}) = t^\gamma f(\tilde{t}) + f(t)$, and the coordinate-wise update rule for these divergences is derived by singling out $q_j$ from the variational bound of forward $f$-divergence VI~\eqref{forward_fbound} introduced in \hyperref[sec_A5]{Section~A.5}. Hence, the $f$-variational bound~\eqref{forward_fbound} can be reformulated as
	\begin{align*}
		\mathcal{L}_f(q_j^{}, q_{-j}^{}, \mathcal{D}) & =  \Ex^{}_q \left[ f\left(\frac{p}{q}\right)  \right] = \Ex^{}_q \left[ f\left(\frac{1}{q} \cdot p \right)  \right] \allowdisplaybreaks \\ 
		& = \mathbb{E}_q \left[ \left( \frac{1}{q_j \cdot q_{-j}} \right)^\gamma \cdot f(p) + f\left( \frac{1}{q_j \cdot q_{-j}}  \right) \right]\\ \allowdisplaybreaks
		& = \mathbb{E}_q \left[ \left(\frac{1}{q_j}\right)^\gamma \left( \frac{1}{q_{-j}} \right)^\gamma f(p) + \left( \frac{1}{q_j} \right)^\gamma f\left(\frac{1}{q_{-j}}\right) + f\left(\frac{1}{q_j}\right)   \right] \allowdisplaybreaks \\ 
		& = \mathbb{E}_{q_j} \left[  \left(\frac{1}{q_j}\right)^\gamma \cdot \mathbb{E}_{q_{-j}} \left[ \left( \frac{1}{q_{-j}} \right)^\gamma f(p) + f\left(\frac{1}{q_{-j}}\right) \right] + f\left(\frac{1}{q_j}\right) \right] \allowdisplaybreaks \\ 
		& = \mathbb{E}_{q_j} \left[  \left(\frac{1}{q_j}\right)^\gamma \cdot f \circ f^{-1} \left( \mathbb{E}_{q_{-j}} \left[ f\left(\frac{p}{q_{-j}}\right) \right]   \right)  + f\left(\frac{1}{q_j}\right) \right]\\
		& = \mathbb{E}_{q_j} \left[  \left(\frac{1}{q_j}\right)^\gamma f(m_j) + f\left( \frac{1}{q_j} \right) \right]\\
		& = \mathbb{E}_{q_j} \left[ f\left( \frac{m_j}{q_j} \right)  \right],
	\end{align*}
	where $m_j = f^{-1}(\mathbb{E}_{q_{-j}}[f(p / q_{-j})])$ can be regarded as an unnormalized probability distribution. After scaling and normalizing $m_j$ into a probability distribution $\tilde{m}_j$ with normalization constant $c > 0$, we have $\mathcal{L}_f(q_j^{}, q_{-j}^{}, \mathcal{D}) = c \cdot D_f(\tilde{m}_j \| q_j)$, which attains its minimum at $\tilde{m}_j = q_j$. Therefore, to minimize the $f$-variational bound when $f \in \mathcal{F}_0$, the marginal distribution $q_j$ should be updated with~\eqref{rule2}:
	\begin{equation}
		q_j \propto m_j = f^{-1}\left( \mathbb{E}_{q^{}_{-j}}\left[  f\left(\frac{p(z, \mathcal{D})}{q^{}_{-j}(z_{-j}^{})}\right)  \right] \right).
	\end{equation}

	\subsection{Mean-field $f$-variational inference under KL divergence}
	For mean-field $f$-VI, we only show an example associated with KL divergence. For KL divergence, consider the $f$-function $f(t) = t\log t \in \mathcal{F}_1$ with $f^*(t) = -\log t$ and $f^{*-1}(t) = \exp(-t)$. Hence, the coordinate-wise update rule~\eqref{rule1} takes the form
	\begin{equation*}
		q^*_j \propto \exp \left(  \Ex_{q^{}_{-j}} \left[ \log p(z, \mathcal{D}) \right] - \mathbb{E}_{q^{}_{-j}}\left[\log q^{}_{-j} \right] \right) \propto  \exp \left(  \Ex_{q^{}_{-j}} \left[ \log p(z, \mathcal{D}) \right]  \right),
	\end{equation*}
	which is in accordance with the update rule of CAVI algorithm~\cite{Bishop_2006}. Demonstrations and experimental results of this update rule can be easily found in the early developments of KL-VI~\cite{Bishop_2006, Murphy_2012, Blei_JASA_2017}. An analytic update rule requires conditionally conjugate models, while some recent advances tried to extend mean-field VI to non-conjugate models~\cite{Knowles_NIPS_2011, Wang_JMLR_2013}. Mean-field $f$-VI subject to other divergences are left to the interested readers to explore.

	\subsection{Mean-field $f$-variational inference algorithm}
	A reference algorithm to implement mean-field $f$-VI is given in the following table.
	
	\setlength{\algoheightrule}{0.75pt} 
	\setlength{\algotitleheightrule}{0.3pt} 
	\begin{algorithm}[H]
		\caption{Mean-field $f$-VI}
		\SetAlgoLined
		\KwIn{Dataset $\mathcal{D} = \{x_n\}_{n=1}^N$,	mean-field variational family $q(z, \theta) = \prod_{j=1}^{J}q_j(z_j, \theta_j)$, model $p(z, x)$, $f$-function $f(\cdot)$, and $f$-variational bound $\mathcal{L}_f(q_\theta^{}, \mathcal{D})$.}
		\renewcommand{\KwIn}{\textbf{Initialize: }} 
		\KwIn{Variational parameters $\theta$ in recognition model $q(z, \theta)$.}\\
		\While{$\mathcal{L}_f(q_\theta^{}, \mathcal{D})$ has not converged}{
			Update parameters $\theta_j$ in $q_j(z_j, \theta_j)$ for $j\in\{1, \cdots,  J\}$ with update rule~\eqref{rule1} or \eqref{rule2}\; 
			Compute $f$-variational bound $\mathcal{L}_f(q_\theta^{}, \mathcal{D})$.}
		\renewcommand{\KwResult}{\textbf{Return: }} 
		\KwResult{Recognition distribution $q(z, \theta)$.}
	\end{algorithm}

	\section{Experiments}
	Detailed descriptions on the experimental settings and the supplementary empirical results are provided in this section.

	\subsection{Synthetic example}\label{secE_1}
\begin{figure}[htpb]
		\centering\vspace{-1.5em}
		\includegraphics[width=0.55\textwidth]{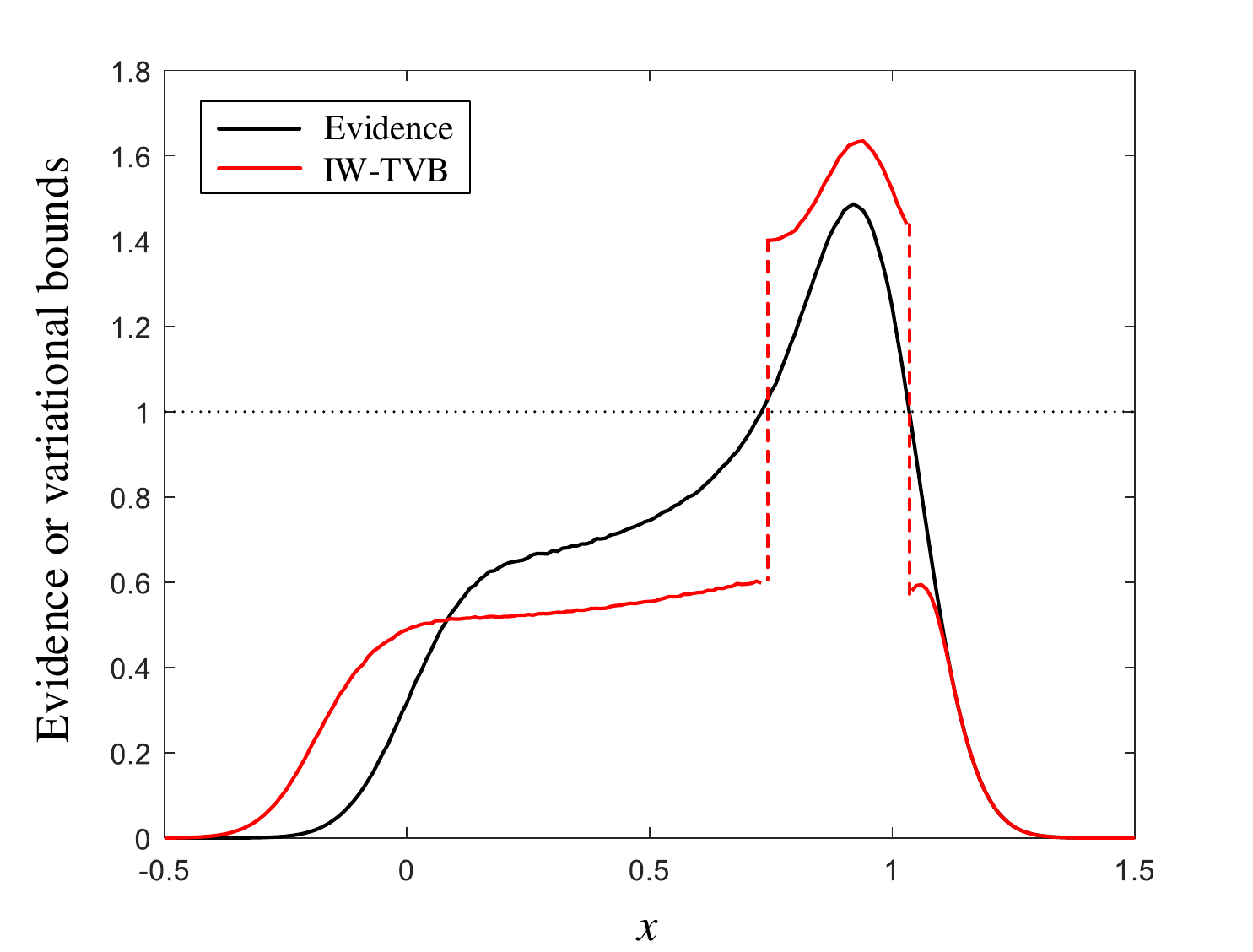}\vspace{-0.5em}
		\caption{Evdience and IW-TVB.}\label{fig3}\vspace{-1em}
	\end{figure}

	For the synthetic example in the original paper, we consider a batch of i.i.d. datapoints generated by the latent variable model $x = \sin(z) + \mathcal{N}(0, 0.01)$, $z \sim \textrm{UNIF}(0, \pi)$. To estimate the true evidence $p(\mathcal{D})$ and $f$-variational bounds, we posit a prior distribution $p(z) = \textrm{UNIF}(0, \pi)$, a likelihood distribution $p(z|x) = \mathcal{N}(\sin(z), 0.01)$, and an approximate model $q^{}_\theta(z) = \textrm{UNIF}(\frac{1-\theta}{2} \pi, \frac{\theta + 1}{2} \pi)$, which is a uniform distribution centered at $z = \pi / 2$ with width $\theta \pi$. The true evidence $p(\mathcal{D})$ is approximated by a naive MC estimator $\hat{p}(x) = \sum_{k=1}^{K}p(x|z_k)$ with $K = 5\times 10^{5}$, and all the other (importance-weighted) $f$-variational bounds in~\hyperref[fig1]{Figure~1} and~\hyperref[fig3]{Figure~3} are estimated by their corresponding MC estimators with $L = 8$ and $K = 5\times 10^4$. Fixing $\theta = 1.1$, we approximate the importance-weighted $f$-variational bound subject to total variation distance (IW-TVB) in~\hyperref[fig3]{Figure~3}, which verifies~\eqref{TVB} and \hyperref[cor1]{Corollary~2}. 
	
	However, it is still worth noting that numerical issues and biased estimators can contaminate the empirical results or cause the violations of theory, despite the fact that the importance-weighted technique can attenuate these flaws by improving the tightness of bounds and their estimators. The estimation of IW-RVB in \hyperref[fig1]{Figure~1} and IW-TVB on $x\in[-0.5, 0]$ in \hyperref[fig3]{Figure~3} are some examples. More discussions and examples on these problems can be found in~\cite{Li_NIPS_2016, Tao_ICML_2018}.

		\subsection{Bayesian neural network}
		Our Bayesian regression framework is developed on the basis of~\cite{Li_NIPS_2016}. The regression model is a single ReLU layer with 50 hidden units for small datasets and 100 hidden units for large datasets (Protein). The likelihood function is selected as $p(y| x, z) = \mathcal{N}(y; F_z(x), \sigma^2)$, where $\sigma$ is a hyper-parameter and $F_z(x)$ is the prediction or output of the neural network with weights $z$. We posit a standard normal prior $z \sim \mathcal{N}(0, I)$ for network weights and a Gaussian approximation $q(z) = \mathcal{N}(\mu^{}_\theta, \textrm{diag}(\sigma_\theta^2))$ to the true posterior, where the variational parameters $\mu^{}_\theta$ and $\sigma_\theta$ are to be optimized. Importance-weighted $f$-variational bounds and their gradients are approximated by MC estimators with $L = 5$, $K = 50$ for small datsets and $K = 10$ for large datasets. Twelve datasets from the UCI Machine Learning Repository~\cite{UCI} are employed, in which six datasets (Boston, CCPP, Concrete, Protein, Wine and Yacht) are the benchmarks previously tested in~\cite{Tao_ICML_2018, Li_NIPS_2016, Wang_NIPS_2018}, while the other six datasets\footnote{(full name, \#instances, \#attributes) of six new benchmarks are provided: \textit{Airfoil} ({Airfoil Self-Noise, 1503, 6}), \textit{Aquatic} (QSAR Aquatic Toxicity, 546, 9), \textit{Building} (Residential Building Data Set, 372, 105), \textit{Fish Toxicity} (QSAR Fish Toxicity, 908, 7), \textit{Real Estate} (Real Estate Valuation Data Set, 414, 7), and \textit{Stock} (Stock Portfolio Performance, 315, 12).} are new benchmarks for VI testing. Each dataset is randomly split into $90\% / 10\%$ for training and testing. The test RMSE and test negative log-likelihood reported in~\hyperref[tab2a]{Table~2} are collected from 20 trials with 500 training epochs in each trial for small datasets and 5 trials with 200 training epochs in each trial for large datasets. For dataset Building, we predict the sale prices and scale the test RMSE by 0.01 for uniform representation. For dataset Stock, we only use the 5-year data to predict the annual return, and the test RMSEs are scaled by 100.
		
		Six $f$-VIs, including three well-established $f$-VIs (KL-VI, R\'enyi's $\alpha$-VI with $\alpha = 3$, and $\chi$-VI with $n = 2$) and three new $f$-VIs (VIs subject to total variation distance and two custom $f$-divergences), are tested and compared in this Bayesian regression example and the following $f$-VAE example. The total variation bound is defined as $\textrm{TVB} = \mathbb{E}_{q(z)}\left[   |p(z, \mathcal{D}) / q(z) - 1|  \right]$, and since $p(z, \mathcal{D}) / q(z) \in (0, 1)$ always holds in \hyperref[sec4_2]{Section~4.2} and \hyperref[sec4_3]{Section~4.3}, we optimize the objective function $\log(\textrm{TVB} - 1)$ for numerical stability. Meanwhile, we also consider i) a custom $f$-VI induced by the dual function $f_{\textrm{c1}}^*(t) = \tilde{f}^*(t) - \tilde{f}^*(1)$, where $\tilde{f}^*(t) = -1/6 \cdot (\log t +t_0)^3 - 1/2 \cdot (\log t + t_0)^2 - (\log t + t_0) - 1$, $t = p(z, \mathcal{D}) / q(z)$, and $t_0 \in \mathbb{R}$ is a parameter to be optimized, and ii) a custom $f$-VI induced by the dual function $f_{\textrm{c2}}^*(t) = \log^2 t + \log t$, which is convex on $t\in (0, 1)$ and can be modified to be a valid $f$-function by reassigning the mapping on $t \in [1, +\infty)$. More feasible $f^*$-functions can be generated from the known $f^*$-functions via the operations that preserve convexity, \textit{e.g.} non-negative weighted sums. While the $f$-VI framework applies to arbitrary valid $f$-functions in theory, the empirical implementations require the $f$-functions and the corresponding estimators to have good numerical properties such that the optimization algorithms can converge. To meet this requirement, we sometimes have to compromise the unbiasedness of estimators, for example, while the CUBO ($n = 2$) employed for regression in~\hyperref[tab2a]{Table~2} should be an upper bound of evidence in theory, the empirical CUBO approximated by a biased estimator in~\cite{Li_NIPS_2016} behaves like a lower bound in the training processes, despite the augmentation of importance-weighted technique.
		
		\vspace{-1em}
		
		\renewcommand{\thetable}{\arabic{table}a}
		\setcounter{table}{1}
		\setlength{\tabcolsep}{2pt}
		\begin{table}[H]\label{tab2a}
			\newcommand\ColWidth{48}
			\caption{Average test error.}\vspace{0.2em}
			\centering	\footnotesize
			\begin{tabular}{ L{47 pt} C{\ColWidth pt} C{\ColWidth pt} C{\ColWidth pt} C{\ColWidth pt} C{\ColWidth pt} C{\ColWidth pt}}
				\toprule[1pt]
				\multirow{2}{*}{Dataset}	& \multicolumn{6}{c}{Test RMSE (lower is better)}\\
				\cmidrule(lr){2-7}  
				& KL-VI & $\chi$-VI  & $\alpha$-VI & TV-VI & $f_{\textrm{c1}}$-VI & $f_{\textrm{c2}}$-VI\\   
				\cmidrule{1-7}  
				Airfoil & \bf{2.16$\pm$.07} & 2.36$\pm$.14 & 2.30$\pm$.08 &  2.47$\pm$.15   & 2.34$\pm$.09 &  2.16$\pm$.09   \\
				Aquatic & \bf{1.12$\pm$.06} & 1.20$\pm$.06 & 1.14$\pm$.07 &  1.23$\pm$.10  & 1.14$\pm$.06 &  1.14$\pm$.06   \\
				Boston &\bf{2.76$\pm$.36}& 2.99$\pm$.37 & 2.86$\pm$.36 &   2.96$\pm$.36   &   2.87$\pm$.36 &  2.89$\pm$.38    \\
				Building & 1.38$\pm$.12 & 2.82$\pm$.51 & 1.83$\pm$.22& 2.57$\pm$.59   & 1.80$\pm$.21&  \bf{1.36$\pm$.15}   \\
				CCPP & \bf{4.05$\pm$.09} & 4.14$\pm$.11 & 4.06$\pm$.08 &   4.19$\pm$.12    &   4.33$\pm$.12&   4.33$\pm$.12  \\
				Concrete &5.40$\pm$.24& \bf{3.32$\pm$.34} & 5.32$\pm$.27& 5.27 $\pm$.24   & 5.26$\pm$.21& 5.32$\pm$.24  \\
				Fish Toxicity & .885$\pm$.037 & .905$\pm$.043 & .891$\pm$.037 &    .878$\pm$.044    &    .883$\pm$.034&   \bf{.862$\pm$.040}  \\
				Protein & 1.93$\pm$.19 & 2.45$\pm$.42 & \bf{1.87$\pm$.17}  & 2.91$\pm$.89     &   1.97$\pm$.21& 1.97$\pm$.20  \\
				Real Estate &  7.48{$\pm$}1.41 & 7.51{$\pm$}1.44 & \bf{7.46{$\pm$}1.42} &   8.02$\pm$1.58   & 7.52$\pm$1.40&  7.99$\pm$1.55   \\
				Stock & 3.85$\pm$1.12 & 3.90$\pm$1.09 & 3.88$\pm$1.13 &      4.33$\pm$.43      & \bf{3.82$\pm$1.11}&   4.18$\pm$.42  \\
				Wine & .642$\pm$.018 & .640$\pm$.021 & .638$\pm$.018 &      .645$\pm$.014      &  .643$\pm$.019&   \bf{.637$\pm$.016}  \\
				Yacht & \bf{0.78$\pm$.12} & 1.18$\pm$.18 & 0.99$\pm$.12 &   1.03$\pm$.14         & 1.00$\pm$.18&  0.82$\pm$.16   \\
				\bottomrule[1pt]
			\end{tabular}
		\end{table}
		
		\vspace{-2em}
		\renewcommand{\thetable}{\arabic{table}b}
		\setcounter{table}{1}
		\setlength{\tabcolsep}{1pt}
		\begin{table}[H]\label{tab2b}
			\newcommand\ColWidth{50}
			\caption{Average negative log-likelihood.}\vspace{0.2em}
			\centering	\footnotesize
			\begin{tabular}{ L{50 pt} C{\ColWidth pt} C{\ColWidth pt} C{\ColWidth pt} C{\ColWidth pt} C{\ColWidth pt} C{\ColWidth pt}}
				\toprule[1pt]
				\multirow{2}{*}{Dataset}	& \multicolumn{6}{c}{Test negative log-likelihood (lower is better)}\\
				\cmidrule(lr){2-7}  
				&   KL-VI & $\chi$-VI  & $\alpha$-VI & TV-VI & $f_{\textrm{c1}}$-VI & $f_{\textrm{c2}}$-VI \\   
				\cmidrule{1-7}  
				Airfoil & \bf{2.17$\pm$.03} & 2.27$\pm$.03 & 2.26$\pm$.02 &  2.28$\pm$.04    & 2.29$\pm$.02   &  2.18$\pm$.03 \\
				Aquatic & \bf{1.54$\pm$.04} & 1.60$\pm$.08 & 1.54$\pm$.07 &  1.56$\pm$.07    &  1.54$\pm$.06  & 1.55$\pm$.04  \\
				Boston & 2.49$\pm$.08 & 2.54$\pm$.18 & \bf{2.48$\pm$.13} &  2.51$\pm$.18     & 2.49$\pm$.13&  2.51$\pm$.10   \\
				Building & 6.62$\pm$.02 & 6.94$\pm$.13 & 6.79$\pm$.03 &    6.88$\pm$.08 & 6.74$\pm$.04& \bf{6.55$\pm$.02}   \\
				CCPP & \bf{2.82$\pm$.02} & 2.84$\pm$.03 & 2.82$\pm$.02 &     2.83$\pm$.02       & 2.95$\pm$.01&  2.91$\pm$.01  \\
				Concrete &3.10$\pm$.04 & \bf{2.61$\pm$.18} & 3.09$\pm$.04&  3.10$\pm$.05   & 3.09$\pm$.03&   3.10$\pm$.04 \\
				Fish Toxicity & 1.28$\pm$.04 & 1.27$\pm$.04 & 1.29$\pm$.04 & \bf{1.26$\pm$.05}  & 1.29$\pm$.03&  1.26$\pm$.03 \\
				Protein & \bf{2.00$\pm$.07} & 2.01$\pm$.08 & 2.04$\pm$.08 &  2.04$\pm$.11 & 2.21$\pm$.04&  2.11$\pm$.05 \\
				Real Estate & 3.60{$\pm$}.30 & 3.70{$\pm$}.45 & \bf{3.59{$\pm$}.32}&   3.86$\pm$.52    & 3.62{$\pm$}.33&  3.74$\pm$.37 \\
				Stock & -1.09$\pm$.04 & -1.09$\pm$.04 & -1.09$\pm$.04 &  -1.73$\pm$.15    & -1.09$\pm$.04&   \bf{-1.84$\pm$.12}    \\
				Wine &.966$\pm$.027 & .965$\pm$.028 & .964$\pm$.025 &  .969$\pm$.023  & .975$\pm$.027&      \bf{.959$\pm$.023}      \\
				Yacht & \bf{1.70$\pm$.02} & 1.79$\pm$.03 & 1.82$\pm$.01 &   1.78$\pm$.02    & 2.05$\pm$.01&  1.86$\pm$.02    \\
				\bottomrule[1pt]
			\end{tabular}
		\end{table}
		
		\vspace{-1.5em}
		
		\subsection{Bayesian variational autoencoder}
\begin{figure}[H]
			\centering
			\includegraphics[width=0.88\textwidth]{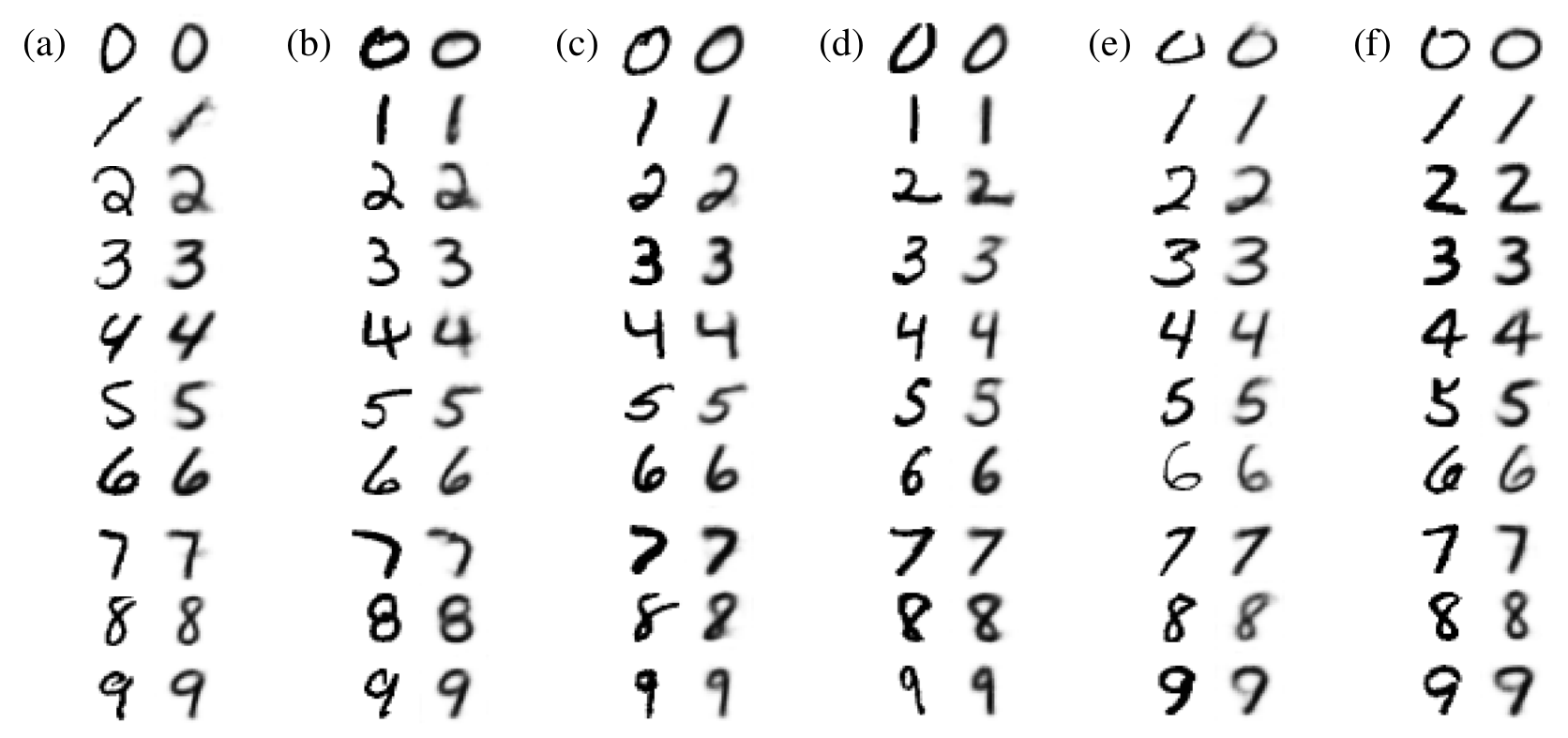}\vspace{-0.2em}
			\caption{Reconstruction of MNIST handwritten digits. Left column shows the original digits. Right column shows the reconstructed digits. (a) is from IW-ELBO loss. (b) is from IW-CUBO ($n = 2$) loss. (c) is from IW-RVB ($\alpha = 3$) loss. (d) is from IW-TVB loss. (e) is from custom $f_{\textrm{c1}}$-variational bound loss, and (f) is from custom $f_{\textrm{c2}}$-variational bound loss.}\label{fig4}
		\end{figure}

		Our Bayesian VAE example is built on the basis of~\cite{MATLAB_VAE}. The encoder network downsamples from a $28\times 28$ or $28\times 20$ image to a 20-dimensional latent space and sequentially consists of i) a $3\times3$ 2-D convolution layer with stride 2, ii) a ReLU layer, iii) a $3\times3$ 2-D convolution layer with stride 2, iv) a ReLU layer, and v) a fully connected layer. The decoder network scales up the 20-dimensional encoding back into a $28\times28$ or $28\times20$ image and sequentially consists of i) a $7\times7$ or $7\times 5$ transposed 2-D convolution layer with stride $[7, 7]$ or $[7, 5]$, ii) a ReLU layer, iii) a $3\times3$ transposed 2-D convolution layer with stride 2, iv) a ReLU layer, v) a $3\times3$ transposed 2-D convolution layer with stride 2, vi) a ReLU layer, and vii) a $3\times3$ transposed 2-D convolution layer. The sizes of training/testing datasets are respectively, $7803/868$, $1768/197$, $60000/10000$, and $24345/8070$ for Caltech 101 Silhouettes, Frey Face, MNIST, and Omniglot, and the mini-batch sizes are respectively $64$, $32$, $512$, and $256$. The loss functions or the importance-weighted $f$-variational bounds are approximated by single-sample MC estimators with $K = 1$ and $L = 3$. After 20 trials with 200 training epochs in each trial, the average test reconstruction errors (lower is better) measured by cross-entropy are given in~\hyperref[tab3]{Table~3}. Some reconstructed and generated images from $f$-VAEs are presented in~\hyperref[fig4]{Figure~4} to \hyperref[fig8]{Figure~8}. While one can improve the quality of these images and reduce the average reconstruction errors in \hyperref[tab3]{Table~3} by adopting more complex encoder and decoder networks, in this experiment, we are more interested in the relative performance of different $f$-VIs.
		
		\vspace{-0.5em}
		
		\vspace{-0.7em}
		
		\begin{figure}[H]
			\centering
			\includegraphics[width=0.95\textwidth]{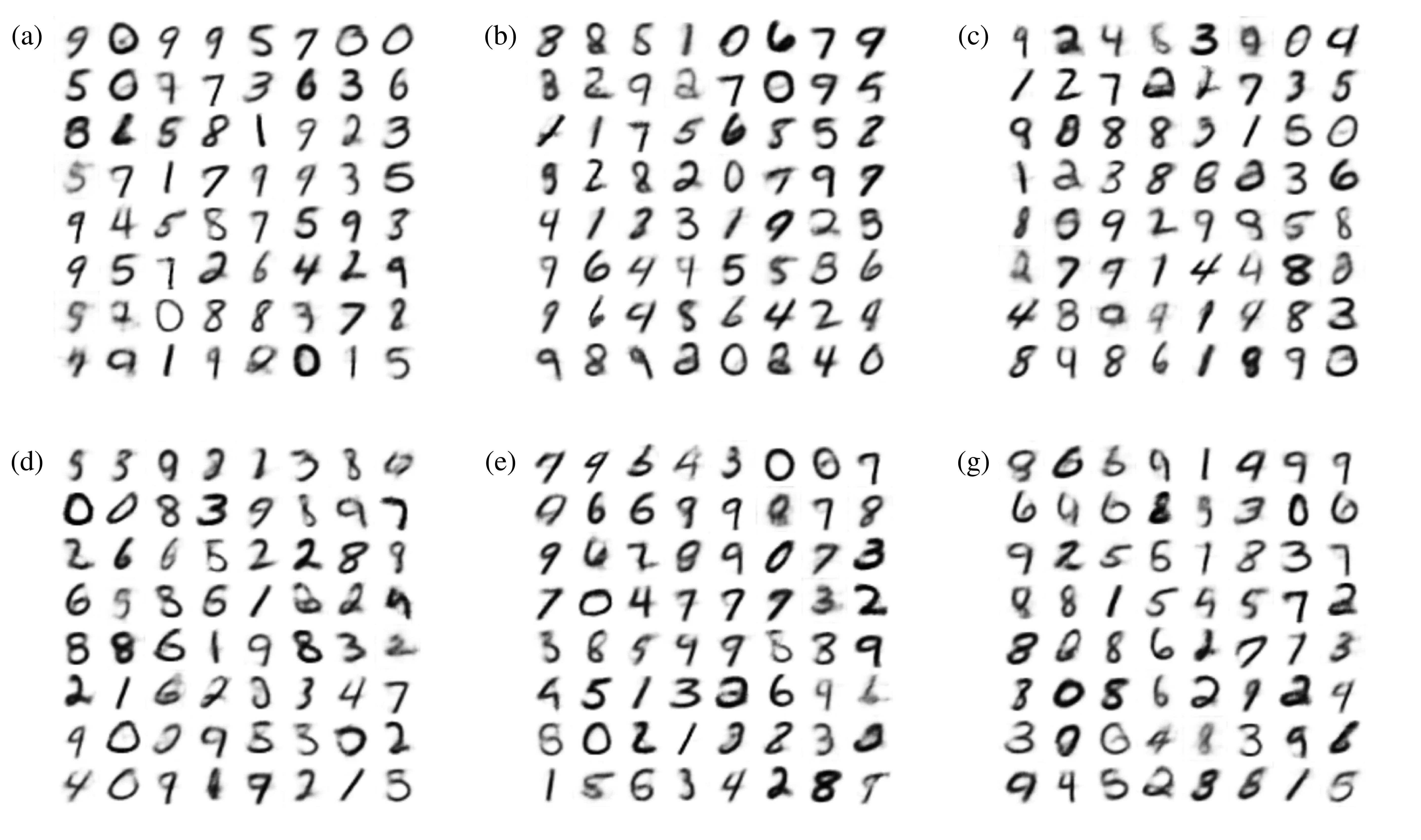}\vspace{-0.5em}
			\caption{Generation of MNIST handwritten digits. (a) is from IW-ELBO loss. (b) is from IW-CUBO ($n = 2$) loss. (c) is from IW-RVB ($\alpha = 3$). (d) is from IW-TVB loss. (e) is from custom $f_{\textrm{c1}}$-variational bound loss, and (f) is from custom $f_{\textrm{c2}}$-variational bound loss.}\label{fig5}
		\end{figure}
		
		\begin{figure}[H]
			\centering
			\includegraphics[width=0.95\textwidth]{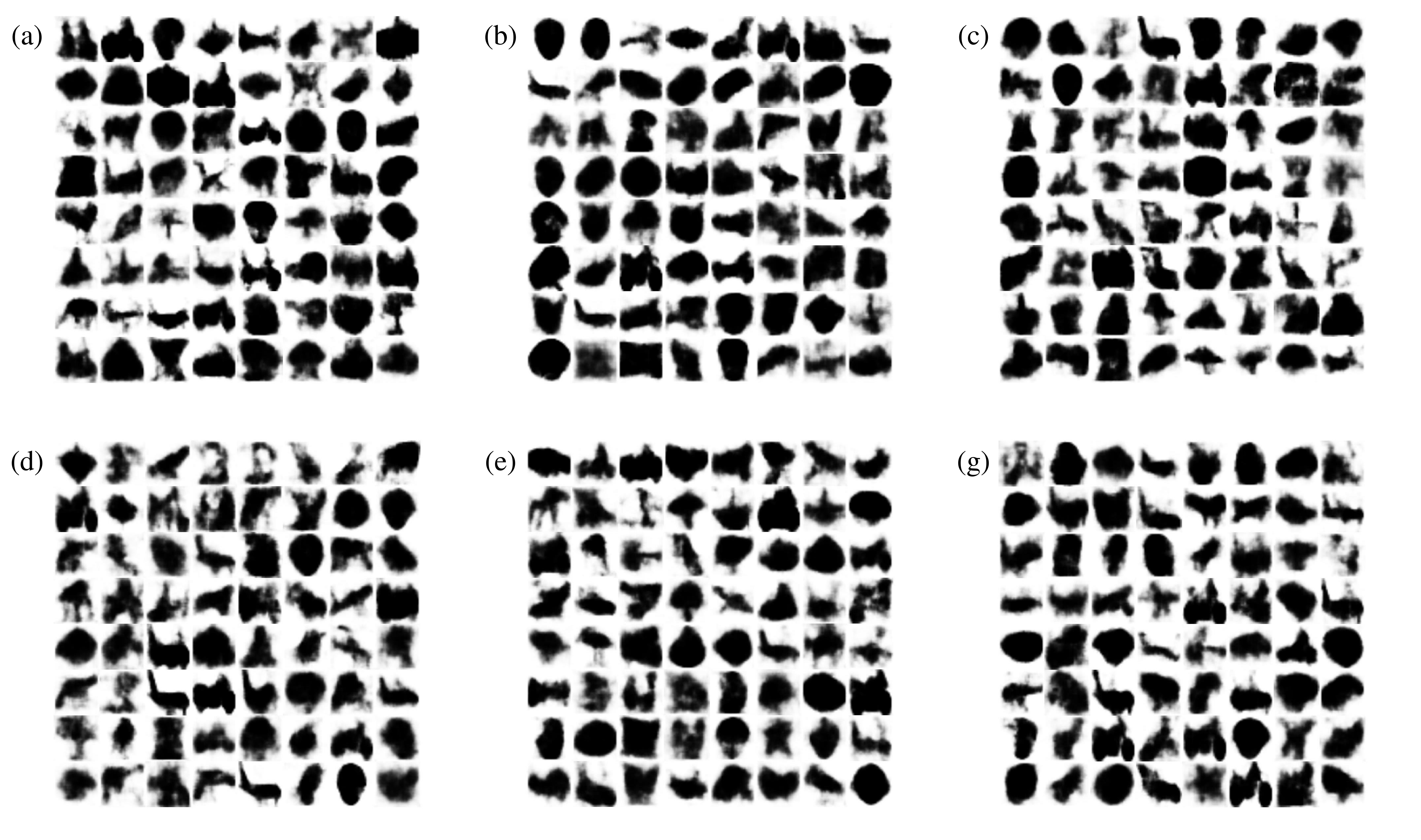}\vspace{-0.5em}
			\caption{Generation of Caltech 101 silhouettes. (a) is from IW-ELBO loss. (b) is from IW-CUBO ($n = 2$) loss. (c) is from IW-RVB ($\alpha = 3$). (d) is from IW-TVB loss. (e) is from custom $f_{\textrm{c1}}$-variational bound loss, and (f) is from custom $f_{\textrm{c2}}$-variational bound loss.}\label{fig6}
		\end{figure}
		
		\begin{figure}[H]
			\centering
			\includegraphics[width=0.97\textwidth]{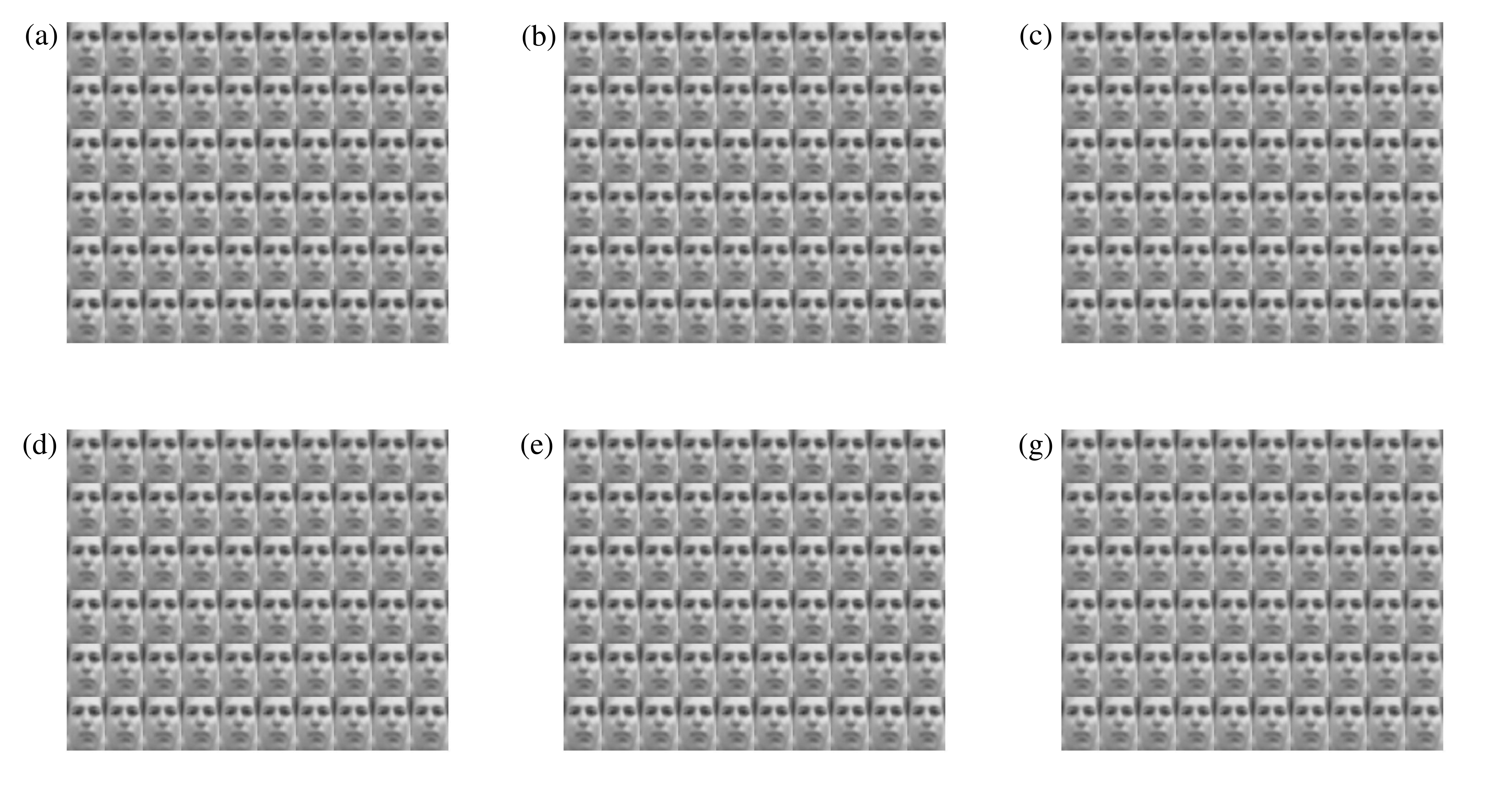}\vspace{-0.5em}
			\caption{Generation of Frey Face. (a) is from IW-ELBO loss. (b) is from IW-CUBO ($n = 2$) loss. (c) is from IW-RVB ($\alpha = 3$). (d) is from IW-TVB loss. (e) is from custom $f_{\textrm{c1}}$-variational bound loss, and (f) is from custom $f_{\textrm{c2}}$-variational bound loss.}\label{fig7}
		\end{figure}
		
		\begin{figure}[H]
			\centering
			\includegraphics[width=0.95\textwidth]{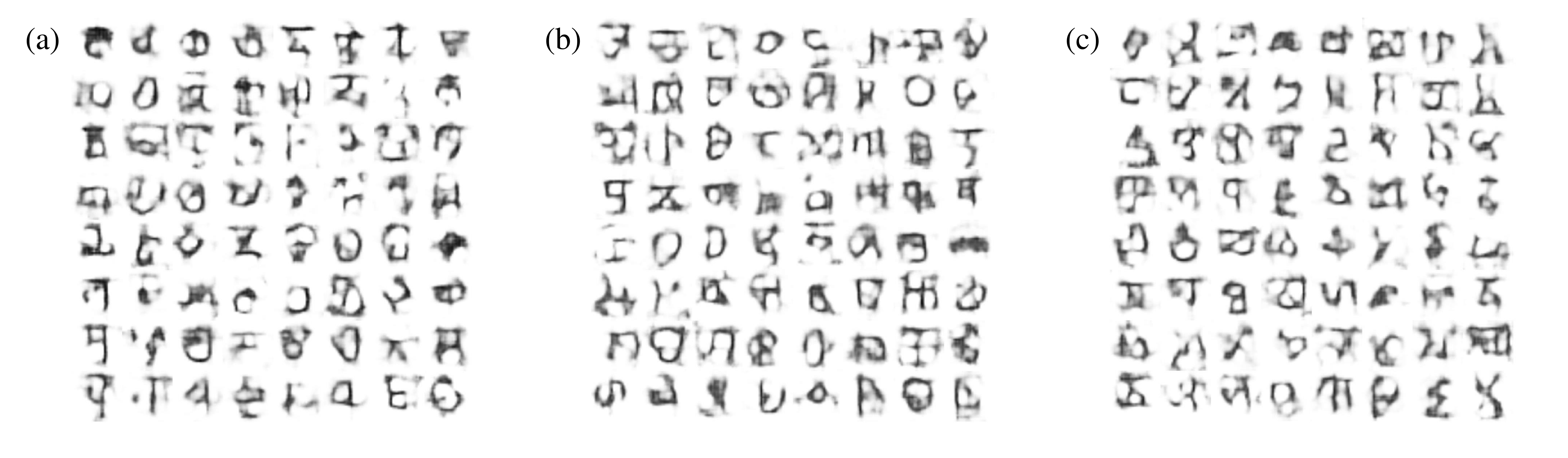}\vspace{-0.5em}
		\end{figure}
		
		\begin{figure}[H]
			\centering
			\includegraphics[width=0.95\textwidth]{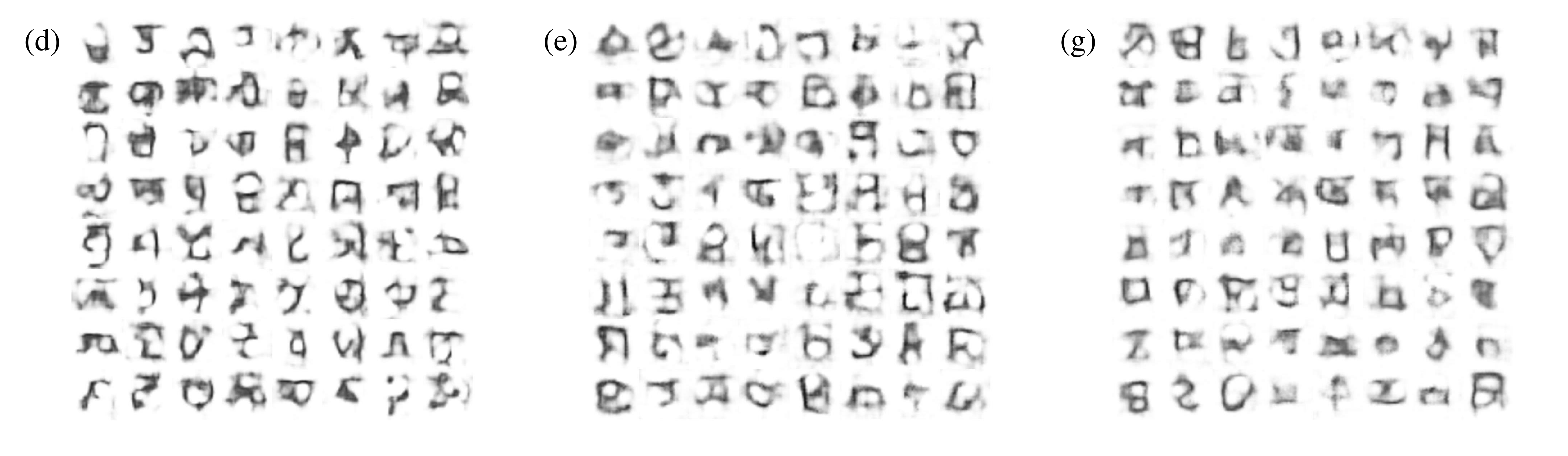}\vspace{-0.5em}
			\caption{Generation of Omniglot alphabets. (a) is from IW-ELBO loss. (b) is from IW-CUBO ($n = 2$) loss. (c) is from IW-RVB ($\alpha = 3$). (d) is from IW-TVB loss. (e) is from custom $f_{\textrm{c1}}$-variational bound loss, and (f) is from custom $f_{\textrm{c2}}$-variational bound loss.}\label{fig8}
		\end{figure}

\end{document}